\documentclass[11pt]{article}

\usepackage{fullpage}
\usepackage[ruled, linesnumbered, vlined, commentsnumbered]{algorithm2e}
\usepackage{amsfonts,amssymb,amsmath,amsthm,amsopn,mathrsfs,mathtools,wasysym}	
\usepackage{hhline,booktabs,colortbl,multirow,tabularx,diagbox,threeparttable} 
\usepackage[caption=false,font=normalsize,labelfont=sf,textfont=sf]{subfig}
\usepackage{graphicx} 
\usepackage{color,xcolor} 
\usepackage{arydshln}
\usepackage{enumerate}
\usepackage{authblk}
\usepackage{footnote}
\usepackage{hyperref}
\usepackage{prettyref}
\usepackage{cite}
\usepackage{setspace}

\usepackage{scrpage2}
\usepackage{geometry}

\usepackage{tikz}
\usepackage{pgfplots}
\usetikzlibrary{positioning,shapes,shadows,arrows,calc}
\tikzstyle{component}=[rectangle, draw=black, rounded corners, fill=blue!40, drop shadow, text centered, anchor=north, text=white, minimum height=1cm]
\tikzstyle{arrow}=[->, thick]

\pgfplotsset{compat=1.12}
\usetikzlibrary{intersections}
\usetikzlibrary{pgfplots.statistics}
\usepgfplotslibrary{fillbetween}

\definecolor{myblue}{RGB}{34,31,217}
\definecolor{mycyan}{gray}{.7}
\definecolor{Gray}{gray}{0.9}

\definecolor{mycyan}{gray}{.7}
\newtheorem{remark}{Remark}

\newtheorem{theorem}{Theorem}

\DeclareMathOperator*{\argmax}{argmax}
\DeclareMathOperator*{\argmin}{argmin}

\newcommand{\pref}{\prettyref}

\newcommand{\bb}[1]{\cellcolor{mycyan}{\textbf{{#1}}}}

\def\our{\texttt{DETO}}

\newrefformat{fig}{Fig.~\ref{#1}}
\newrefformat{tab}{Table~\ref{#1}}
\newrefformat{sec}{Section~\ref{#1}}
\newrefformat{alg}{Algorithm~\ref{#1}}
\newrefformat{property}{Property~\ref{#1}}
\newrefformat{theorem}{Theorem~\ref{#1}}
\newrefformat{definition}{Definition~\ref{#1}}
\newrefformat{corollary}{Corollary~\ref{#1}}
\newrefformat{lemma}{Lemma~\ref{#1}}
\newrefformat{conj}{Conjecture~\ref{#1}}
\newrefformat{def}{Definition~\ref{#1}}
\newrefformat{eq}{equation~(\ref{#1})}
\newrefformat{app}{Appendix~\ref{#1}}

\usepackage{lscape}

\begin{document}

\title{\vspace{-1ex}\LARGE\textbf{A Data-Driven Evolutionary Transfer Optimization for Expensive Problems in Dynamic Environments}~\footnote{This manuscript is submitted for potential publication. Reviewers can use this version in peer review.}}

\author[1]{\normalsize Ke Li}
\author[2]{\normalsize Renzhi Chen}
\author[2]{\normalsize Xin Yao}
\affil[1]{\normalsize Department of Computer Science, University of Exeter, EX4 4QF, Exeter, UK}
\affil[2]{\normalsize National Innovation Institute of Defence Technology, Beijing, China}
\affil[3]{\normalsize Department of Computer Science and Engineering, Southern University of Science and Technology, Shenzhen 518055, China}
\affil[$\ast$]{\normalsize Email: \texttt{k.li@exeter.ac.uk}}

\date{}
\maketitle

\vspace{-3ex}
{\normalsize\textbf{Abstract: } }    Many real-world problems are usually computationally costly and the objective functions evolve over time. Data-driven, a.k.a. surrogate-assisted, evolutionary optimization has been recognized as an effective approach for tackling expensive black-box optimization problems in a static environment whereas it has rarely been studied under dynamic environments. This paper proposes a simple but effective transfer learning framework to empower data-driven evolutionary optimization to solve dynamic optimization problems. Specifically, it applies a hierarchical multi-output Gaussian process to capture the correlation between data collected from different time steps with a linearly increased number of hyperparameters. Furthermore, an adaptive source task selection along with a bespoke warm staring initialization mechanisms are proposed to better leverage the knowledge extracted from previous optimization exercises. By doing so, the data-driven evolutionary optimization can jump start the optimization in the new environment with a strictly limited computational budget. Experiments on synthetic benchmark test problems and a real-world case study demonstrate the effectiveness of our proposed algorithm against nine state-of-the-art peer algorithms.



\section{Introduction}
\label{sec:introduction}

Real-world black-box optimization problems are complex and challenging not only because they are non-convex, multi-modal, highly constrained, multi-objective, but also subject to various uncertainties. In particular, dynamic optimization problem (DOP), where the objective function(s) experience abrupt changes over time with temporally changing/evolving optima, is ubiquitous in many real-world optimization scenarios. For example, modern software systems are usually configurable with a set of known control features (e.g., CPU cap, thread pool size and cache size) that affect the system's functional and non-functional properties. To adapt to the requirements of many end-users concurrently, those control features need to be configured dynamically on-the-fly according to the changing runtime environments, e.g., bandwidth and response time~\cite{ChenLY18}.

Partially due to the population-based property and the self-adaptivity~\cite{FogelM04}, evolutionary algorithms (EAs) have been widely recognized as an effective tool for DOPs. Most EA developments implicitly assume that the objective function evaluations (FEs) are computationally cheap or trivial. Thus, it is common to see an EA routine costs at least tens or hundreds thousands or even more FEs in a single run. Unfortunately, such assumption usually does not stand in real-world optimization scenarios where FEs are physically, computationally or economically expensive, as known as expensive black-box optimization. It becomes more challenging when the environments dynamically change with time and are full of uncertainties.

The Gaussian process (GP) model~\cite{RasmussenW06}, also known as Kriging~\cite{Stein99} or design and analysis of computer experiments (DACE) stochastic process model~\cite{SantnerWN18}, has been widely recognized as one of the most efficient and effective tools for dealing with expensive black-box optimization problems~\cite{Jones01,Frazier18,ShahriariSWAF16}. Its basic assumption is built upon a GP from which each testing point is sampled. The distribution of any unknown point is estimated based on the data collected from the previous search. The update of the GP model is guided by the optimization of an acquisition function, e.g., expected improvement~\cite{Jones01}, probability of improvement~\cite{Frazier18} and upper confidence bound~\cite{SrinivasKKS10}, that implements a principled exploitation versus exploration balance over the surrogate landscape. In practice, the GP model constitutes the foundation of data-driven optimization approaches, including but not limited to the efficient global optimization (EGO)~\cite{JonesSW98} and the Bayesian optimization (BO)~\cite{ShahriariSWAF16}, for expensive black-box optimization problems. There also have been many efforts devoted to the use of GP and other machine learning models under the banner of EAs, as known as surrogate-assisted EAs (SAEAs)~\cite{JinWCGM19,LiC22}. Surprisingly, the existing data-driven optimization endeavors have been extensively dedicated for the continuous space with static objective function(s), whereas few works (only~\cite{Morales-EncisoB15,nyikosa2018bayesian,RichterSCRL20,FanLT20} to the best of our knowledge) have been done for DOPs. This can be attributed to the following two imperative challenges.
\begin{itemize}
    \item The ultimate goal of dynamic optimization is no longer merely approximating the global optimum, but to track the evolving optimum over time. Due to the dynamic environments and the fast moving problem landscapes, it is almost impossible or at least ineffective and inefficient to restart, also known as \lq cold start\rq, the optimization routine from scratch when the environment changes, e.g., the first four strategies proposed in~\cite{Morales-EncisoB15}.
    \item Although most data-driven optimization approaches have the ability to adapt to the underlying search landscapes, it is not wise to simply ignore the changes but just let the optimization routine run. The time-varying nature of the DOP itself renders the data collected in previous time steps become less reliable or even harmful for model building in the current time step, e.g.,~\cite{Morales-EncisoB15,nyikosa2018bayesian,RichterSCRL20,Chen021a}. Furthermore, excessive data turn the training and inference of a GP model into a computationally expensive practice.
\end{itemize}

Note that DOPs implicitly assume certain correlations between problem landscapes of recent time steps, given that many  applications usually experience a gentle change from one stage to the next. Therefore, it is reasonable to selectively leverage the data collected from the previous search processes to \lq warm start\rq\ the underlying optimization task. In fact, warm staring a BO has been studied in the machine learning community such as hyperparameter optimization~\cite{LindauerH18} and neural architecture search~\cite{Hu0CMHD19}. However, they are mainly a \lq one-off\rq\ process that aims to transfer knowledge/experience learned from the existing data to \lq jump start\rq\ the underlying problem-solving process. This is in principle a static version of our dynamic optimization scenario from one time step to the next in particular whereas the problem-solving tasks and problem landscapes continuously change over time.

Very recently, we have developed a simple BO variant~\cite{Chen021a}, dubbed \texttt{TBO}, that uses a multi-output GP (MOGP)~\cite{AlvarezRL12} to exploit and leverage the correlations between different time steps, instead of modeling each of which individually. The proof-of-concept results have demonstrated the effectiveness of \texttt{TBO} for solving DOPs in comparison to some peer algorithms with a limited computational budget. However, it is far from mature in view of the following three drawbacks.
\begin{itemize}
    \item The number of hyperparameters of the conventional MOGP is cubic of the number of tasks. In our dynamic optimization scenario, it is highly unlikely to collect abundant data at each time step since each FE is costly. In this case, the increase of data can hardly match the overflow of the number of hyperprameters. This can make the MOGP to be underfit with the time step goes by.
    \item In \texttt{TBO}, it applies a simple decay mechanism to proactively dump some previously collected data in case they are less relevant to the current time step, to mitigate the notoriously time-consuming of MOGP. However, since this decay mechanism is augmented as a hyperparameter of MOGP, it not only aggravates the risk of overfitting given the limited training data in expensive optimization, but also does not provide a controllable method to pick up the most relevant source for the MOGP model training.
    \item Last but not the least, one of the advantages of \texttt{TBO} is that it only needs to sample a limited number of initial solutions to jump start the new optimization task by MOGP. However, since this sampling at the initialization step is conducted in a random manner, it can easily be biased and the knowledge about the previous optimization tasks cannot be fully exploited.
\end{itemize}

Bearing these concerns in mind, this paper proposes a principled data-driven transfer optimization framework, dubbed \our, for expensive DOPs. The main contributions of this paper are outlined as follows.
\begin{itemize}
    \item To mitigate the underfitting problem of the conventional MOGP, we propose a hierarchical MOGP that replaces the block matrices of the task related hyperparameters of the MOGP with a series of masked block matrices.
    \item To mitigate the risk of overfitting caused by the augmented hyperparameter(s), we develop an adaptive source data selection mechanism to proactively pick up the most suitable source task for MOGP model training.
    \item To enforce a strategic jump start at the outset of a new optimization task, we propose a bespoke initialization strategy by leveraging the local optima collected from the estimated problem landscapes of the previously finished optimization tasks, each of which is represented as a GP models respectively.
    \item To overcome the non-convex property of optimizing the acquisition function, we propose a simple but effective EA with local search to search for the next point of merit.
    \item Extensive experiments on both synthetic benchmark problems with controllable dynamic characteristics and a real-world dynamic optimization problem fully demonstrate the effectiveness our proposed \our\ in comparison to other nine peer algorithms.
\end{itemize}

The rest of this paper is organized as follows. \pref{sec:preliminary} provides some important background knowledge in order to understand this paper along with a pragmatic literature review. The technical details of our proposed \our\ are delineated in~\pref{sec:method} and its performance is validated through a series of empirical studies given in~\pref{sec:experiments}. At the end, \pref{sec:conclusions} concludes this paper and shed some lights on future directions.


\section{Preliminaries}
\label{sec:preliminary}

This section starts with some preliminary knowledge including the formal definition of DOP considered in this paper and the basic idea of a vanilla data-driven optimization method based on the GP model. Then, we give a pragmatic literature review on both data-driven methods (EGO and BO variants in particular) and EAs for solving DOPs.

\subsection{Problem Definition}
\label{sec:problem_definition}

The DOP considered in this paper is defined as:
\begin{equation}
    \begin{aligned}
        \mathrm{maximize} &\; f(\mathbf{x},t) \\
        \mathrm{subject\ to}     &\; \mathbf{x} \in \Omega
    \end{aligned},
\end{equation}
where $\mathbf{x}=(x_1,\cdots,x_n)^\top$ is a decision vector (variable), $\Omega=[x_i^L,x_i^U]_{i=1}^n\subset\mathbb{R}^n$ is the search space, $t\in\{1,\cdots,T\}$ is a discrete time step and $T>1$ is the total number of time steps. Note that the objective function is assumed to be \textit{computationally expensive} and it changes over time so as its landscapes and the local/global optima. In this case, the ultimate goal of computationally expensive black-box optimization in dynamic environments is to track the time-varying optimum across $T$ time steps under a strictly limited amount of overall computational budget.

\begin{theorem}(Karush-Kuhn-Tucker conditions~\cite{Bertsekas16})
    If $\mathbf{x}^\ast$ is a local optimum of an objective function $f(\mathbf{x}):\mathbb{R}^n\rightarrow\mathbb{R}$ with $m$ constraints $\{g_i(\mathbf{x})\leq 0\}_{i=1}^m$ and the set of vectors $\{\nabla g_j(\mathbf{x}^\ast)|j\ \text{is the index of an active constraint}\}$ are linearly independent. There exists a vector $\mu=(\mu_1,\cdots,\mu_m)^\top\in\mathbb{R}^m$, such that:
    \begin{equation}
        \begin{aligned}
            \nabla f(\mathbf{x}^\ast)+\sum_{j=1}^m\mu_j\nabla g_j(\mathbf{x}^\ast)&=0\\
            \mu_jg_j(\mathbf{x}^\ast)|_{j=1}^m&=0
        \end{aligned},
        \label{eq:kkt}
    \end{equation}
    where $\mu_i\geq 0$, $\forall i\in\{1,\cdots,m\}$.
    \label{theorem:kkt}
\end{theorem}

\begin{remark}
    The objective and constraint functions are assumed to be continuously differentiable in the Karush-Kuhn-Tucker (KKT) conditions.
\end{remark}

\begin{remark}
    The objective function considered in this paper does not include constraints, thus we ignore the $\sum_{j=1}^m\mu_j\nabla g_j(\mathbf{x}^\ast)$ part of~\pref{eq:kkt} in the latter derivations.
\end{remark}

\subsection{Vanilla Data-Driven Optimization Based on GP Model}
\label{sec:bo}

Data-driven optimization based on GP model is a class of sequential model-based optimization methods for solving black-box optimization problems with computationally expensive objective function(s). As shown in the line $1$ of~\pref{alg:bo}, it starts from uniformly sampled solutions according to a space-filling experimental design (e.g., Latin hypercube sampling). Thereafter, it sequentially updates its next sample until the computational budget is exhausted.

\begin{algorithm}[t!]
    \KwIn{Related hyperparameter settings.}
    \KwOut{The best solution found so far.}
    Use an experimental design method to sample a set of initial solutions $\mathcal{X}\leftarrow\{\mathbf{x}^i\}_{i=1}^{N_I}$ from $\Omega$ and evaluate their objective function values $\mathcal{Y}\leftarrow\{f(\mathbf{x}^i)\}_{i=1}^{N_I}$. Set the initial training dataset $\mathcal{D}\leftarrow\{\langle\mathbf{x}^i,f(\mathbf{x}^i)\rangle\}_{i=1}^{N_I}$;\\
\While{termination criteria is not met}{
    Build a GP model based on $\mathcal{D}$;\\
    Optimize an acquisition function to obtain a candidate solution $\hat{\mathbf{x}}^\ast$;\\
    Evaluate the objective function values of $\hat{\mathbf{x}}^\ast$ and update $\mathcal{D}\leftarrow\mathcal{D}\bigcup\{\langle\hat{\mathbf{x}}^\ast,f(\hat{\mathbf{x}}^\ast)\rangle\}$;
}
    \Return $\argmin\limits_{\mathbf{x}\in\mathcal{D}}f(\mathbf{x})$
    \caption{Pseudo code for a vanilla data-driven optimization based on GP model}
    \label{alg:bo}
\end{algorithm}

The data-driven optimization based on GP model consists of two main components: $i$) a surrogate model based on GP for approximating the expensive objective function; and $ii$) an infill criterion (based on the optimization of an acquisition function) that decides the next sampling point for being evaluated by the actual expensive objective function. We briefly introduce these two components as follows.

\subsubsection{Surrogate Model}
\label{sec:GP}

This paper considers using GP regression (GPR) model to serve the surrogate modeling (line $3$ of~\pref{alg:bo}). Given a set of training data $\mathcal{D}=\left\{\langle\mathbf{x}^i,f(\mathbf{x}^i)\rangle\right\}_{i=1}^{N}$, the GPR model considered in this paper aims to learn a noise-free latent function with a constant $0$ mean. For each testing input vector $\mathbf{z}^\ast\in\Omega$, the mean and variance of the target $f(\mathbf{z}^\ast)$ are predicted as:
\begin{equation}
    \begin{aligned}
        \mu(\mathbf{z}^\ast)&={\mathbf{k}^\ast}^TK^{-1}\mathbf{f}\\
        \sigma(\mathbf{z}^\ast)&=k(\mathbf{z}^\ast,\mathbf{z}^\ast)-{\mathbf{k}^\ast}^T K^{-1} {\mathbf{k}^\ast}
    \end{aligned},
    \label{eq:GP}
\end{equation}
where $X=(\mathbf{x}^1,\cdots,\mathbf{x}^N)^\top$ and $\mathbf{f}=(f(\mathbf{x}^1),\cdots,f(\mathbf{x}^N))^\top$. $\mathbf{m}(X)$ is the mean vector of $X$, $\mathbf{k}^\ast$ is the covariance vector between $X$ and $\mathbf{z}^\ast$, and $K$ is the covariance matrix of $X$. In particular, a covariance function, also known as a kernel function, is used to measure the \textit{similarity} between a pair of two data points $\mathbf{x}$ and $\mathbf{x}^\prime\in\Omega$. Here we use the radial basis function (RBF) kernel in this paper and it is defined as:
\begin{equation}
    k(\mathbf{x},\mathbf{x}^\prime)=\gamma\exp(-\frac{\|\mathbf{x}-\mathbf{x}^\prime\|}{\ell})^2,
\end{equation}
with two hyperparameters $\gamma$ and length scale $\ell$. The predicted mean $\mu(\mathbf{z}^\ast)$ is directly used as the prediction of $f(\mathbf{z}^\ast)$, and the predicted variance $\sigma(\mathbf{z}^\ast)$ quantifies the uncertainty. As recommended in~\cite{RasmussenW06}, the hyperparameters associated with the mean and covariance functions are learned by maximizing the log marginal likelihood function defined as:
\begin{equation}
    \log p(\mathbf{f}|X)=-\frac{1}{2}\mathbf{f}^\top K^{-1}\mathbf{f}-\frac{1}{2}\log|K|-\frac{N}{2}\log 2\pi.
    \label{eq:likelihood}
\end{equation}

\subsubsection{Infill Criterion}
\label{sec:infill}

Instead of optimizing the surrogate objective function, the search for the next point of merit $\hat{\mathbf{x}}^\ast$ is driven by an acquisition function that strikes a balance between exploitation of the predicted minimum and exploration of model uncertainty:
\begin{equation}
    \hat{\mathbf{x}}^\ast=\argmax_{\mathbf{x}\in\Omega}f^{\mathrm{acq}}(\mathbf{x}),
    \label{eq:infill}
\end{equation}
where $f^{\mathrm{acq}}(\mathbf{x})=\mu(\mathbf{x})+\omega\sigma(\mathbf{x})$ is the widely used upper confidence bound (UCB)~\cite{Jones01} in this paper. In particular, $\omega>0$ is a parameter that controls the trade-off between exploration and exploitation. The optimization of (\ref{eq:infill}) can be carried out by either an EA or a gradient descent method (lines $4$ and $5$ of~\pref{alg:bo}).

\subsection{Related Works}
\label{sec:related}

This subsection gives a pragmatic overview of related works for handling DOPs from the data-driven and evolutionary optimization perspectives respectively.

\subsubsection{Data-Driven Optimization}
\label{sec:ddo}

As briefly introduced in Section I, there is few prior works on expensive dynamic optimization from the data-driven optimization perspective. To the best of our knowledge, Kushner~\cite{Kushner62} is the first who proposed to model a dynamically changing function by using a stochastic processes model. However, only one dimensional problem was considered in~\cite{Kushner62} which is far from useful in practical scenarios. Afterwards, this topic has been kept silent and lied down for years until Morales-Enciso and Branke proposed eight strategies to handle expensive DOPs under the EGO framework~\cite{Morales-EncisoB15}. In particular, the first four simple and straightforward strategies including random sampling, evolutionary mutation, restart and ignorance; while the latter four are proposed to reuse and transfer previous knowledge to the underlying problem-solving process. Obviously, these are too simple to be effective for handling complex dynamics. Although the other four strategies proposed in~\cite{Morales-EncisoB15} aim to selectively leverage the previous information by augmenting the covariance function in GP model building, they are subject to some prior knowledge/assumption of the characteristics of the underlying problem. Recently, Nyikosa et al.~\cite{nyikosa2018bayesian} proposed an adaptive BO that uses a product kernel to augment a spatial-temporal covariance function for handling expensive DOPs. The temporal part is used as a forgetting factor that helps adaptively choose different old data in the GP model training and inference. This is implemented by a time-related length-scale parameter that can be learned as an additional hyperparameter in a maximum likelihood estimation. In~\cite{RichterSCRL20}, two approaches were proposed to enable BO for solving DOPs. One is called window approach that only uses the most recent observations for GP model training; and the other is called time-as-covariate approach that augments the time into the covariance function. Very recently, we proposed a BO variant~\cite{Chen021a} that provides a proof-of-concept result of using a MOGP to leverage the correlation between the current optimization task with the previous ones. To mitigate the soaring computational cost, a simple sliding window strategy is applied to merely focus on the most recent time steps.

Another related line of research is \lq warm starting\rq\ a data-driven optimization by using additional information besides the FEs. Especially in the BO community, a large body of literature promotes the idea to meta-learn models~\cite{PerroneJSA18,FlennerhagMLD19} or acquisition function~\cite{WistubaSS18,VolppFFDFHD20}, to transfer knowledge by weighted aggregating surrogate models across various tasks~\cite{WistubaSS16,SchillingWS16,LindauerH18}, and to conduct multi-tasking~\cite{SwerskySA13}. These methods have shown to be powerful for hyper-parameter optimization, but they usually require large amounts of historical training data to extract the characteristics of a class of tasks, which might hardly be met in the expensive dynamic optimization scenario considered in this paper.

\subsubsection{Evolutionary Optimization}

Due to the population-based and adaptive characteristics, EAs have been widely recognized as a powerful tool for DOPs. However, since EAs usually require tons of FEs to obtain a reasonable solution, it is not directly applicable and computationally unaffordable for expensive DOPs. In recent decades, there have been many efforts devoted to using EAs for solving DOPs, here we just give a pragmatic overview of some selected works according to the ways of handling dynamics. Interested readers are recommended to some excellent survey papers for a more comprehensive literature review, e.g.,~\cite{JinB05,NguyenYB12,MavrovouniotisL17,YazdaniCYBJY21,YazdaniCYBJY21a}.

\begin{itemize}
    \item\textit{\underline{Diversity enhancement}}: As its name suggests, the basic idea is to propel an already converged population jump out of the approximated optimum in the previous time step by increasing the population diversity. This is usually implemented by either responding to the dynamics merely after detecting a change of the environment or maintaining the population diversity throughout the whole optimization process. The prior one is known as diversity introduction and is usually accompanied by a change detection technique. For example, \cite{CobbG93,DebNK06,Nguyen11} developed some hyper-mutation methods by which the solutions are expected to jump out of their current positions and thus to adapt to the changed environment. In~\cite{HuE02,JansonM06,GouveaA15}, some additional solutions, either generated in a random manner or by some heuristics, are injected into the current population to participate the survival competition. On the other hand, the latter one, called diversity maintenance, usually does not explicitly respond to the changing environment~\cite{LiKCLZS12,MavrovouniotisY15,RuanYZZY17} whereas it mainly relies on its intrinsic diversity maintenance strategy to adapt to the dynamics. In addition to diversity introduction and maintenance, a multi-population approach is another alternative to handle dynamics by enhancing diversity. In particular, it mainly maintains multiple sub-populations concurrently, each of which is either responsible for a particular area in the search space~\cite{Fernandez-MarquezA10,ChengY10} or assigned with a specific task such as searching for the global optimum or tracking the dynamic environments~\cite{LiY12a,LiNYYZ15,LiNYMY16}.

    \item\textit{\underline{Memory mechanism}}: The major purpose of this type of method is to utilize the historical information to accelerate the convergence progress. In particular, the memory mechanism implicitly assumes that the DOPs are periodical or recurrent. In other words, the optima may return to the regions near their previous locations. In~\cite{Etaner-UyarH05}, researchers used redundant coding via diploid genomes as the implicit memory to encode history into the solution representation. In contrast, memory can also be maintained explicitly. For example, \cite{ZhaoS09,PengGY11,MavrovouniotisY16} directly stored the previous promising solutions or local optima visited in the previous time steps to inform the current search process. \cite{PengZZ14} exploited the historical knowledge and used it to build up an evolutionary environment model. Accordingly, this model is used to guide the adaptation to the changing environment.

    \item\textit{\underline{Prediction strategy}}:
Since it is reasonable to anticipate some patterns associated with the changes in dynamic environments, the prediction strategy is usually coupled with a memory mechanism to reinitialize the population according to the learned patterns extracted from previous search experience. For example, \cite{HatzakisW06} employed an autoregressive model, built upon the historical information, to seed a new population within the neighborhood of the anticipated position of the new optimal solution. By utilizing the regularity property of the Pareto set of continuous multi-objective optimization problems, \cite{ZhouJZ14} and~\cite{ZhangYJWL20} developed various models to capture the movement of the Pareto set manifolds’ centers under different dynamic environments. Instead of anticipating the position of the new optima when the environment changes, \cite{ZouLYBZ17} suggested to predict the optimal moving direction of the population. More recently, \cite{RossiAD08} and~\cite{MurugananthamTV16} proposed a hybrid scheme, which uses the Kalman Filter or random re-initialization, alternatively, according to their online performance, to respond to the changing environment. 
\end{itemize}


\section{Proposed Algorithm}
\label{sec:method}

The pseudo code of \our\ is given in~\pref{alg:tbo2} which shares almost the same logic as the vanilla data-driven optimization based on GP model shown in~\pref{alg:bo}. The main differences (highlighted in \colorbox{lightgray}{light gray} background) are: $i$) a hierarchical MOGP with a reduced number of hyperparameters as the surrogate model to leverage the information collected from the previous time steps; $ii$) an adaptive source data selection mechanism to mitigate the risk of overfitting; $iii$) a bespoke initialization mechanism that warm starts the new optimization task; and $iv$) a simple but effective EA variant with local search to search for the next point of merit.

\begin{algorithm}[t!]
	\caption{Pseudo code of \our}
	\label{alg:tbo2}
    \KwIn{Related hyperparameter settings.}
    \KwOut{The optima found at all time steps $\mathcal{X}^\ast$.}
	$t\leftarrow 1$, $\mathscr{D}\leftarrow\emptyset$, $\mathcal{X}^\ast\leftarrow\emptyset$\;
    Use an experimental design method to sample a set of initial solutions $\mathcal{X}^1\leftarrow\{(\mathbf{x}^i,t)\}_{i=1}^{N_I}$ from $\Omega$ and evaluate their objective function values $\mathcal{Y}\leftarrow\left\{f((\mathbf{x}^i,t),t)\right\}_{i=1}^{N_I}$. Set the initial training dataset $\mathcal{D}^t\leftarrow\left\{\langle(\mathbf{x}^i,t),f((\mathbf{x}^i,t),t)\rangle\right\}_{i=1}^{N_I}$, $N_t\leftarrow N_I$\;
	\While{$t\leq T$}
	{
    	\While{$N_t<N^\mathrm{FE}_t$}
    	{
            \vbox{\colorbox{lightgray}{\vbox{
    	    Build a MOGP model based on $\mathscr{A}\bigcup\mathcal{D}^t$ with a low-rank approximation of covariance matrix\;
	    	Optimize an acquisition function to obtain a candidate solution $(\hat{\mathbf{x}},t)$\;}}}
            Evaluate the objective function values of $(\hat{\mathbf{x}}^\ast,t)$ and set $\mathcal{D}^t\leftarrow\mathcal{D}^t\bigcup\{\langle(\hat{\mathbf{x}}^\ast,t),f((\hat{\mathbf{x}}^\ast,t),t)\rangle\}$\;
    		$N_t\leftarrow N_t+1$\;
    	}
        $\mathscr{D}\leftarrow\mathscr{D}\bigcup\mathcal{D}^t$\;
        $\mathcal{X}^\ast\leftarrow\mathcal{X}^\ast\bigcup\bigg\{\argmin\limits_{(\mathbf{x},t)\in\mathcal{D}^t}f((\mathbf{x},t),t)\bigg\}$\;
        \vbox{\colorbox{lightgray}{\vbox{
    	$t\leftarrow t+1$\;
        Use the adaptive source data selection to pick up $k$ related datasets from $\mathscr{D}$\;
        Use the warm staring initialization to generate $k$ augmented datasets to construct $\mathscr{A}$\;
		$N_t\leftarrow 0$, $\mathcal{D}^t\leftarrow\emptyset$\;
        }}}
    }
	\Return $\mathcal{X}^\ast$
\end{algorithm}

\subsection{Surrogate Model Building}
\label{sec:mogp}


Given a DOP, we denote $\mathscr{D}=\{\mathcal{D}^1,\cdots,\mathcal{D}^T\}$ as a set of observations collected so far where $\mathcal{D}^t=\left\{((\mathbf{x}^k,t),f((\mathbf{x}^k,t),t)\right\}_{k=1}^{N_t}$, $t\in\{1,\cdots,T\}$, $T\geq 1$ and $N_t>1$ is the number of observations evaluated at the $t$-th time step. This paper proposes to apply a MOGP model to exploit the correlations between the observations collected at different time steps. Comparing to the conventional GPR model, the MOGP model is featured with the multi-output covariance function, also known as linear model of coregionalization (LMC)~\cite{AlvarezRL12}, defined as:
\begin{equation}
    k((\mathbf{x},t),(\mathbf{x}^\prime,t^\prime))=\sum_{i=1}^T[A_i]_{t,t'}k_i(\mathbf{x},\mathbf{x}^\prime), 
    \label{eq:lmc}
\end{equation}
where $(\mathbf{x},t)$ is a candidate solution sampled at the current time step $t$ while $(\mathbf{x}^\prime,t^\prime)$ is an observation at the $t^\prime$-th time step ($1\leq t^\prime\leq T$) chosen from $\mathscr{D}$. $k_i(\ast,\ast)$ is the covariance function at the $i$-th time step ($1\leq i\leq T$) as in the conventional GPR. $A_i=B_iB_i^\top\subset\mathbb{R}^{T\times T}$ is used to characterize the correlation between different time steps where
\begin{equation}
    B_i = 
    \begin{bmatrix}
        b_{1,1} & b_{1,2} & \cdots & b_{1,r} \\
        b_{2,1} & b_{2,2} & \cdots & b_{2,r} \\
        \vdots  & \vdots  & \ddots & \vdots  \\
        b_{T,1} & b_{T,2} & \cdots & b_{T,r} 
    \end{bmatrix},
    \label{eq:coef_matrix}
\end{equation}
where $b_{j,k}$ ($1\leq j\leq T$ and $1\leq k\leq r$) is the coefficient (as the hyperparameter) that aggregates the covariance between outputs across different time steps. For each testing input vector sampled at the $t$-th time step $(\mathbf{z}^\ast,t)$, the mean and variance of the target $f((\mathbf{z}^\ast,t),t)$ are predicted as:
\begin{equation}
    \begin{aligned}
        \mu((\mathbf{z}^\ast,t))&={\mathbf{k}_\mathrm{T}^\ast}^TK^{-1}_\mathrm{T}\mathbf{f}_\mathrm{T}\\
        \sigma((\mathbf{z}^\ast,t))&=k(\mathbf(\mathbf{z}^\ast,t),(\mathbf{z}^\ast,t))-{\mathbf{k}_\mathrm{T}^\ast}^T K^{-1} _\mathrm{T}{\mathbf{k}^\ast}_\mathrm{T}
    \end{aligned},
    \label{eq:mogp}
\end{equation}
where $\mathbf{f}_\mathrm{T}$ is a vector of the objective values of all solutions in $\mathscr{D}$, $\mathbf{k}_\mathrm{T}^\ast$ is the covariance vector between $(\mathbf{z}^\ast,t)$ and solutions in $\mathscr{D}$, and $K_\mathrm{T}$ is the covariance matrix of solutions in $\mathscr{D}$.


According to the equations~(\ref{eq:lmc}) and~(\ref{eq:coef_matrix}), we can infer that the number of hyperparameters of the LMC is $\mathcal{O}(T^3)$, which grows cubically with the number of time steps. Given that the FEs considered in this paper are computationally expensive, we do not expect to collect abundant data across different time steps. In this case, we have the risk of underfitting the LMC in our expensive dynamic optimization scenario. To mitigate this issue, inspired by the hierarchical GP developed in~\cite{ParkC10}, we propose a hierarchical MOGP (HMOGP) which keeps the covariance function the same as~\pref{eq:lmc} but the task correlation matrix $A_i$ is defined as:
\begin{equation}
    A_i =
    \begin{dcases}
        1  & \text{if } t\geq i, t^\prime\geq i\\
        0  & \text{otherwise}
    \end{dcases}.
\end{equation}
Alternatively, we can write $A_i$ as:
\begin{equation}
    A_i = \left[ \begin{matrix}\mathbf{0}_{(i-1)\times (i-1)} & \mathbf{1}_{(i-1)\times (T-i+1)}  \\ \mathbf{1}_{(T-i+1) \times (i-1)} & \mathbf{1}_{(T-i+1) \times (T-i+1)}\end{matrix}\right].
\end{equation}
More specifically, we have:
\begin{equation}
\begin{aligned}
    &A_1=\left[
        \begin{matrix}
            1 & \cdots & 1  \\ \vdots & \ddots & \vdots \\ 1 & \cdots & 1
        \end{matrix}
        \right],
    A_2=\left[
        \begin{matrix}
            0 & 1 & \cdots & 1  \\ 1 & 1 & \cdots & 1 \\ \vdots & \vdots & \ddots & \vdots \\ 1 & 1 &  \cdots & 1
        \end{matrix}
        \right], \\ &\cdots,
    A_T=\left[
        \begin{matrix}
            0 & \cdots & 0 & 1  \\ \vdots & \ddots & \vdots & \vdots \\ 0 & \cdots & 0 & 1 \\ 1 & \cdots & 1 & 1
        \end{matrix}
        \right].
\end{aligned}
\end{equation}
By doing so, the number of hyperparameters of this HMOGP is significantly reduced to two magnitudes of order as $\mathcal{O}(T)$, comparing against the conventional LMC.

\subsection{Adaptive Source Data Selection Mechanism}
\label{sec:source}

\begin{figure}[t!]
    \centering
    \includegraphics[width=.7\linewidth]{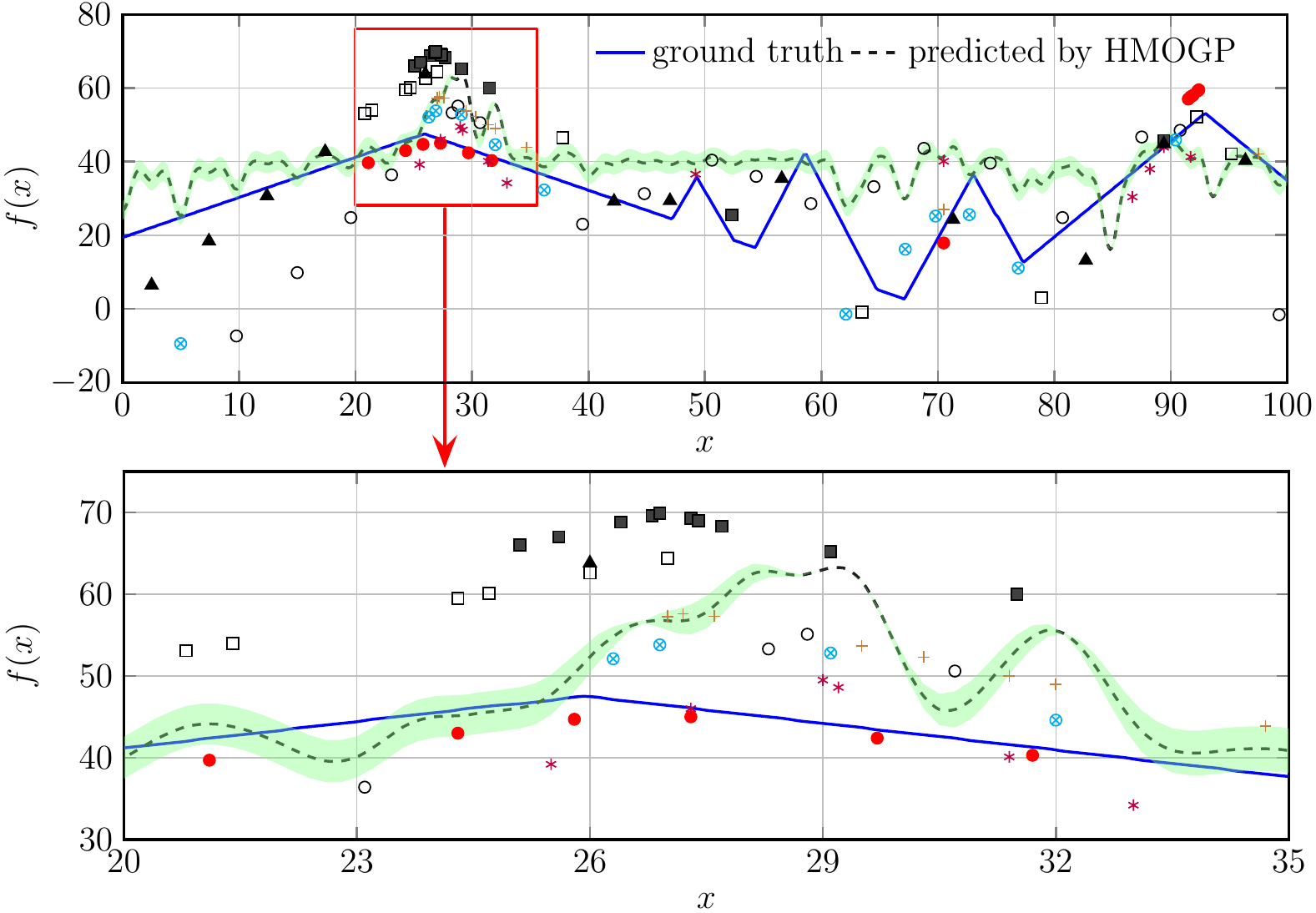}
    \caption{An illustrative example of an overfitting problem in HMOGP. Here we consider source tasks collected from the moving peak benchmark problem at $8$ different time steps, which are represented as different marks.}
    \label{fig:overfit_example}
\end{figure}

During the process of dynamic optimization, we can expect the evaluated solutions are progressively accumulated with time. However, although it sounds to be counter-intuitive at first glance since the data collected during the expensive optimization is usually too limited to train an adequate model, accumulating an \lq excessive\rq\ amount of data can lead to the risk of overfitting. In~\pref{fig:overfit_example}, we use the moving peak function considered in our empirical study in~\pref{sec:setup} to constitute an illustrative example. More specifically, we consider $T=8$ time steps, each of which consists of $19$ data points sampled from a uniform distribution within the range of $\mathbf{x}$ of the moving peak function. As the example shown in~\pref{fig:overfit_example}, it is clear to see that the HMOGP becomes overfit with many unnecessary local optima by considering all data collected from $8$ different source datasets. To mitigate this issue, we propose an adaptive source data selection mechanism to proactively pick up the most suitable previously collected source data to build the HMOGP at each time step. It contains four steps.
\begin{enumerate}[Step 1:]
    \item Build a GPR model $\mathcal{GP}(\mathcal{D}^t)$, $t\in\{1,\cdots,T-1\}$, for each of the previous time steps.
    \item Store the hyperparameters $\mathbf{h}^{t}$ of $\mathcal{GP}(\mathcal{D}^t)$ into $\mathcal{H}=\{\mathbf{h}^t\}_{t=1}^{T-1}$.
    \item Use the classic $k$-means clustering algorithm to divide $\mathcal{H}$ into $k>1$ clusters.
    \item For each cluster, pick up the hyperparameters closest to its centroid and the corresponding dataset is chosen as a source data.
\end{enumerate}

\begin{figure}[t!]
    \centering
    \includegraphics[width=.4\linewidth]{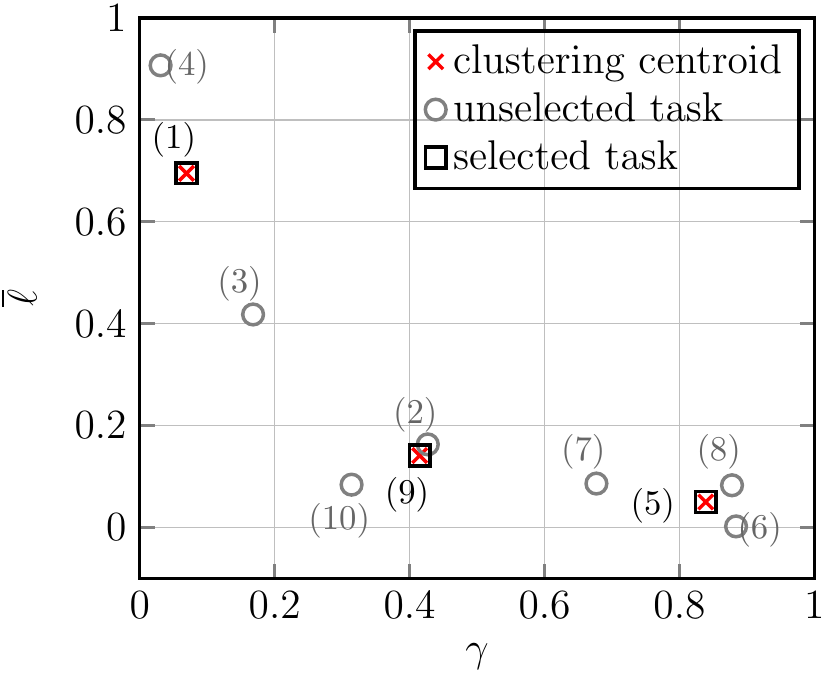}
    \caption{An illustrative example of the adaptive source data selection w.r.t. the hyperparameters of the RBF kernel in HGP for the moving peak problem at $10$ different time steps. Here we set $k=3$ in the $k$-means clustering. The selected tasks are denoted as the black $\square$ symbol while the unselected tasks are denoted as the \textcolor{gray}{gray $\fullmoon$} symbol.}
    \label{fig:task_selection}
\end{figure}

\begin{figure*}[t!]
    \centering
    \includegraphics[width=\linewidth]{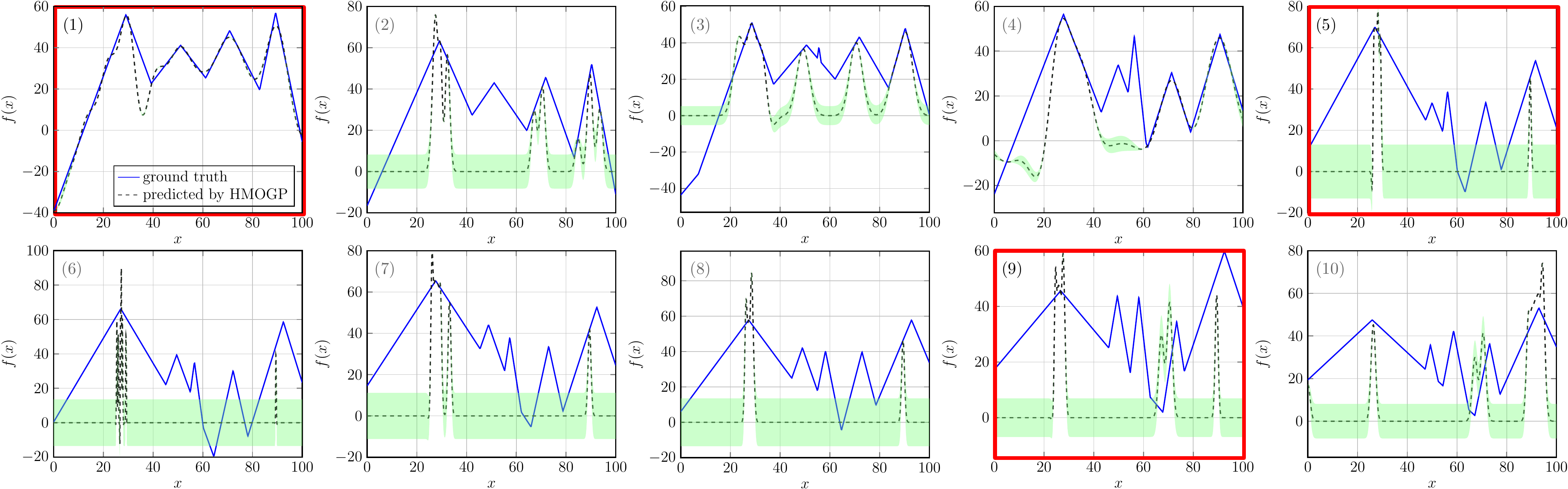}
    \caption{An illustrative example of the fitness landscapes (including both the ground truth versus the prediction made by the corresponding HMOGP) of the moving peak problems at $10$ different time steps. The selected tasks are highlighted with the \textcolor{red}{\textbf{red bold frames}}.}
    \label{fig:landscape_example}
\end{figure*}

\begin{remark}
    In this paper, we use the hyperparameters of $\mathcal{GP}(\mathcal{D}^t)$ as the implicit features to represent the characteristics of the problem landscape of $f(\mathbf{x},t)$ where $t\in\{1,\cdots,T-1\}$.
\end{remark}

\begin{remark}
    In practice, it is not necessary to build all $T-1$ GPR models from scratch in Step 1 and Step 2 each time. Instead, we only need to build the GPR model for the previous time step at the outset of each new optimization task.
\end{remark}

\begin{remark}
    To facilitate the knowledge transfer and to mitigate the potential negative transfer, a common idea for source task selection, widely studied in the transfer learning and multi-task learning literature~\cite{ZhuangQDXZZXH21}, is to identify the most similar task(s). However, in the context of dynamic optimization, we can hardly obtain enough data at the outset of the new time step. This thus largely leads to some unexpected bias if we still try to align the previous tasks with the underlying one. Bearing this consideration in mind, the basic idea of our proposed adaptive source data selection is to find the most representative tasks to serve our purpose.
\end{remark}

Let us use the example in~\pref{fig:task_selection} and~\pref{fig:landscape_example} to illustrate this process. As shown in~\pref{fig:task_selection}, by using the $k$-means clustering algorithm, the hyperparameters (note that $\overline{\ell}$ is normalized $\ell$ within the range $[0,1]$) of $10$ GPR models built by different source datasets are grouped into $3$ clusters. By cross referencing to the fitness landscapes estimated by the corresponding GPR models in~\pref{fig:landscape_example}, we find that $\{\mathcal{D}^1,\mathcal{D}^3,\mathcal{D}^4\}$, $\{\mathcal{D}^5,\mathcal{D}^6,\mathcal{D}^7,\mathcal{D}^8\}$ and $\{\mathcal{D}^2,\mathcal{D}^9,\mathcal{D}^{10}\}$ bear much resemblance to each other within the same cluster whereas they complement the other tasks lying in a different cluster.

\subsection{Warm Starting Initialization Mechanism}
\label{sec:initialization}

In the conventional BO and SAEA, the initialization step is usually implemented by an experimental design method that randomly samples a set of solutions in a strategic manner. However, given the potential recurrent patterns in DOP across different time steps, there should be some useful knowledge collected from the previous optimization tasks. It is counter-intuitive if such knowledge is ignored but just start the new problem-solving process from scratch. This is especially wasteful when considering an expensive optimization scenario in which the computational budget at each time step is highly limited. Bearing this consideration in mind, we develop a bespoke initialization mechanism that leverages the information of related previous tasks recommended by the adaptive source task selection to warm start the underlying optimization task. Specifically, it has the following two major steps.
\begin{enumerate}[Step 1:]
    \item For each of the $k$ tasks recommended by the adaptive source task selection discussed in~\pref{sec:source}, identify the set of local optima $\mathcal{L}^i$ where $i\in\{1,\cdots,k\}$ w.r.t. the corresponding GPR model.
    \item For each task $i\in\{1,\cdots,k\}$, sort the samples in $\mathcal{L}^i$ based on the predicted objective function values. Use the following three-step procedure to pick up $1\leq\sigma<|\mathcal{L}^i|$ samples to constitute the augmented dataset $\mathcal{A}^i$ (started from an empty set) for the $i$-th task.
        \begin{enumerate}[Step 2.1:]
            \item Pick up the current best sample and calculate its Euclidean distance(s) w.r.t. all samples in $\mathcal{A}^i$.
            \item If any of the distances is less than a threshold $\epsilon_\ell$, remove this sample from $\mathcal{L}^i$ and we move to the next sample then go to Step 2.1. Otherwise, this sample is removed from $\mathcal{L}^i$ and it is added into $\mathcal{A}^i$.
            \item Repeat Step 2.1 and Step 2.2 until the size of $\mathcal{A}^i$ reaches $\sigma$.
        \end{enumerate}
\end{enumerate}

\begin{remark}
    Since the GPR model is continuously differentiable, we can derive the first order derivative of the predicted mean function w.r.t. a solution $\mathbf{x}$ as:
    \begin{equation}
        \frac{\partial\overline{g}(\mathbf{x})}{\partial\mathbf{x}}=\frac{\partial\mathbf{k}^\ast}{\partial\mathbf{x}}K^{-1}\mathbf{f}.
        \label{eq:first_order}
    \end{equation}
    Based on the KKT conditions in~\pref{theorem:kkt} and~\pref{eq:first_order}, we can derive the saddle points of the corresponding GPR model by identifying the solutions that satisfy $\frac{\partial\mathbf{k}^\ast}{\partial\mathbf{x}}K^{-1}\mathbf{f}=0$. They are thereafter used as the local optima in Step 1.
\end{remark}

\begin{remark}
    Instead of re-evaluating the objective function values of the selected samples in $\mathcal{A}^i$ where $i\in\{1,\cdots,k\}$, we only use the corresponding GPR model to assign a pseudo objective function value to each sample.
\end{remark}

\begin{remark}
    In Step 2, a na\"ive idea is to pick the top $\sigma$ samples from $\mathcal{L}^i$ where $i\in\{1,\cdots,k\}$. However, it can be less representative if the selected local optima are crowded within a small niche. Thus, we try to enforce certain level of diversity by making sure that the inter distance of samples in $\mathcal{A}^i$ is at least larger than $\epsilon_\ell=10^{-2}\times\max\limits_{1\leq i\leq n} (x_i^U-x_i^L)$.
\end{remark}

\subsection{Optimization of Acquisition Function by a Hybrid EA}
\label{sec:hea}

As discussed in~\cite{WilsonHD18}, the optimization of UCB is non-trivial given its non-convex properties. In view of the outstanding performance of differential evolution (DE)~\cite{StornP97} for global optimization, here we propose a simple but effective DE variant with local search to serve a proof-of-concept purpose. Specifically, it consists of the following six steps.
\begin{enumerate}[Step 1:]
    \item Initialize a population of solutions $\mathcal{P}=\{\mathbf{x}^i\}_{i=1}^N$.
    \item Use DE to generate a population of offspring $\mathcal{Q}$ and let $\overline{\mathcal{P}}=\mathcal{P}\bigcup\mathcal{Q}$.
    \item Pick up the best $\kappa$ candidates from $\overline{\mathcal{P}}$, according to the UCB value, to constitute an archive $\overline{\mathcal{A}}$.
    \item For each candidate $\tilde{\mathbf{x}}\in\overline{\mathcal{A}}$ do
    \begin{enumerate}[Step 4.1:]
        \item Use stochastic gradient descent (SGD)~\cite{GoodfellowBC16} on $\tilde{\mathbf{x}}$ to obtain $\tilde{\mathbf{x}}^\ast$, then set $\overline{\mathcal{P}}=\overline{\mathcal{P}}\cup\{\tilde{\mathbf{x}}^\ast\}\setminus\{\tilde{\mathbf{x}}\}$.
        \item If $\|\tilde{\mathbf{x}}^\ast-\tilde{\mathbf{x}}\|_2<\epsilon_d$, then set $\kappa=\max(\kappa-1,1)$; otherwise, set $\kappa=\min(\kappa+1,2N)$.
    \end{enumerate}
\item Pick up the best $N$ solutions from $\overline{\mathcal{P}}$ to constitute $\mathcal{P}$ according to their UCB values.
\item If the termination criterion is met, then stop and output $(\mathbf{x}^\ast,t)=\underset{\mathbf{x}\in\mathcal{P}}{\argmax}f^{\mathrm{acq}}(\mathbf{x})$. Otherwise, go to Step 2.
\end{enumerate}

\begin{remark}
    This DE variant leverages the global search ability of the DE and the local search characteristics of the SGD to achieve a balance between exploration and exploitation.
\end{remark}

\begin{remark}
    $\kappa$ is a prescribed hyperparameter that controls the scale of the local search. In this paper, we set $\kappa=5$ in the initialization, i.e., Step 1, while it will be adaptively tuned in Step 4.2.
\end{remark}

\begin{remark}
    In Step 4.2, if $\tilde{\mathbf{x}}^\ast$ is very similar to $\tilde{\mathbf{x}}$, i.e., their Euclidean distance (denoted as $\|\cdot\|_2$) is smaller than a threshold $\epsilon_d$, it implies the local search might not be helpful any longer. Therefore, $\kappa$ will be reduced in the next round.
\end{remark}


\section{Experimental Setup}
\label{sec:setup}

In this section, we introduce the experimental settings for validating the effectiveness of our proposed \our\ compared against nine peer algorithms.

\subsection{Benchmark Test Problems}
\label{sec:problems}

In our experiments, we consider two synthetic benchmark problems. One is the moving peaks benchmark (MPB)~\cite{Branke99}, the most popular benchmark test suite in the dynamic optimization literature. MPB generates a landscape that consists of several components. Each component has a peak whose height, width and location change over time with the environment changes. MPB is flexible and can generate scalable functions with a configurable number of components, each of which is able to become the global optimum at the current time step. In addition, we develop another synthetic benchmark problem based on the MPB, named MPB with Gaussian peaks (MPBG). It is featured with Gaussian peaks while the other characteristics are kept the same. The mathematical definitions and visual interpretations of the fitness landscapes of both MPB and MPBG can be found in the Appendix A of the supplemental document of this paper\footnote{The supplemental document is located in~\url{https://tinyurl.com/c2nd64ar}.}.

\subsection{Peer Algorithms}
\label{sec:parameters}

The differences of nine selected peer algorithms versus our proposed \our\ are outlined as follows.
\begin{itemize}
    \item\underline{Restart BO~(\texttt{RBO})}: This is the vanilla BO that restarts from scratch when the environment changes.
    \item\underline{Continuous BO~(\texttt{CBO})}: Different from \texttt{RBO}, it ignores the time-varying nature of DOP but applies all data collected so far in $\mathscr{D}$ to train a GP model. In view of the cubic complexity of training and inference of GP, \texttt{CBO} can be excessively time consuming with the increase of time steps in our dynamic optimization setting. To mitigate this issue, we only consider the most recent five time steps in our experiments.

    \item\underline{Discounted information via noise sampling~(\texttt{DIN})}~\cite{Morales-EncisoB15}: This is best variant reported in~\cite{Morales-EncisoB15} for tracking the varying global optimum in dynamic environments. It augments a noise term as a function of the age of the samples into the covariance function of GP.

    \item\underline{Model-based optimization~(\texttt{MBO})}~\cite{RichterSCRL20}: Instead of using the previously collected data in model training, \texttt{MBO} proposed an augmented acquisition function to search for the next point of merit.

    \item\underline{Time varying GP bandit optimization~(\texttt{TV-GP-UCB})}~\cite{BogunovicSC16}: It is an extension of the classical GP upper confidence bound algorithm~\cite{SrinivasKKS10}. To dump old and irrelevant data in a smooth fashion, it aggregates a decaying factor into a spatial-temporal kernel.

    \item\underline{BO with box mechanism~(\texttt{Box+GP})}~\cite{PerroneS19}: Instead of searching from a set of arbitrarily defined search ranges, it automatically designs the BO search space geometries by relying on previous FEs. This strategy endows the existing BO methods with transfer learning properties.

    \item\underline{Ranking-weighted GP ensemble~(\texttt{RGPE})}~\cite{FeurerLHB18}: It develops an ensemble model for BO by aggregating GPs built from previously collected data as a linear combination. In particular, the weights are learned in a transfer learning setting based on agnostic Bayesian learning of ensembles.

    \item\underline{Boosted hierarchical GP~(\texttt{BHGP})}~\cite{TighineanuSBRBV22}: By using an ad-hoc independent assumption between the source and target data, it develops a boosted version of hierarchical GP to speed up BO from a transfer learning perspective.
    
    \item\underline{Transfer BO~(\texttt{TBO})}~\cite{Chen021a}: It uses a MOGP to leverage the information collected from the previous time steps. In addition, a simple decay mechanism is applied to downgrade the impact of \lq older\rq\ samples collected from the previous time steps.
\end{itemize}

\subsection{Performance Metrics}
\label{sec:metrics}

Since the landscape and optimum change over time in DOP, it is not adequate to simply evaluate the quality of the best solution found by an algorithm in performance comparison. Instead, we need to take more information (e.g., how the algorithms track the moving optima) into account. In this paper, we consider the following two performance metrics.
\begin{itemize}
    \item\underline{Average error overtime}: it measures the average deviation from the global optimum over all FEs:
        \begin{equation}
            \bar{\epsilon}_\mathrm{f}=\frac{1}{N_\mathrm{FE}}\sum_{i=1}^{N_\mathrm{FE}}\Big[f(\mathbf{x}^\ast,t)-f_i(\bar{\mathbf{x}}^\ast,t)\Big],
        \end{equation}
        where $N_\mathrm{FE}$ is the total number of FEs, $f(\mathbf{x}^\ast,t)$ is the global optimum at the $t$-th time step, $t\in\{1,\cdots,T\}$, and $f_i(\bar{\mathbf{x}}^\ast,t)$ is the best solution found at the $i$-th FE of the $t$-th time step.

    \item\underline{Average error for each time step}: it measures the average accuracy achieved over all time steps:
        \begin{equation}
            \bar{\epsilon}_\mathrm{t}=\frac{1}{T}\sum_{t=1}^{T}\Big[f(\mathbf{x}^\ast,t)-f(\hat{\mathbf{x}}^\ast,t)\Big],
        \end{equation}
        where $f(\mathbf{x}^\ast,t)$ is the true global optimum at the $t$-th time step, $t\in\{1,\cdots,T\}$, and $f(\hat{\mathbf{x}}^\ast,t)$ is the best solution found at the end of the $t$-th time step.

%
\end{itemize}

\subsection{Statistical Tests}
\label{sec:stats}

To have a statistical interpretation of the significance of comparison results, we use the following three statistical measures in our empirical study.
\begin{itemize}
	\item\underline{Wilcoxon signed-rank test}~\cite{Wilcoxon1945IndividualCB}: This is a non-parametric statistical test that makes little assumption about the underlying distribution of the data and has been recommended in many empirical studies in the EA community~\cite{DerracGMH11}. In particular, the significance level is set to $p=0.05$ in our experiments.

    \item\underline{Scott-Knott test}~\cite{MittasA13}: Instead of merely comparing the raw metric values, we apply the Scott-Knott test to rank the performance of different peer techniques over 31 runs on each test problem. In a nutshell, the Scott-Knott test uses a statistical test and effect size to divide the performance of peer algorithms into several clusters. The performance of peer algorithms within the same cluster are statistically equivalent. Note that the clustering process terminates until no split can be made. Finally, each cluster can be assigned a rank according to the mean metric values achieved by the peer algorithms within the cluster. In particular, the smaller the rank is, the better performance of the algorithm achieves.

    \item\underline{$A_{12}$ effect size}~\cite{VarghaD00}: To ensure the resulted differences are not generated from a trivial effect, we apply $A_{12}$ as the effect size measure to evaluate the probability that one algorithm is better than another. Specifically, given a pair of peer algorithms, $A_{12}=0.5$ means they are \textit{equivalent}. $A_{12}>0.5$ denotes that one is better for more than 50\% of the times. $0.56\leq A_{12}<0.64$ indicates a \textit{small} effect size while $0.64 \leq A_{12} < 0.71$ and $A_{12} \geq 0.71$ mean a \textit{medium} and a \textit{large} effect size, respectively. 
\end{itemize}
Note that both Wilcoxon signed-rank test and $A_{12}$ effect size are also used in the Scott-Knott test for generating clusters.

\subsection{Parameter Settings}
\label{sec:parameters}

The parameter settings used in our experiments are summarized as follows.
\begin{itemize}
    \item\underline{Test problem settings}: We consider different number of variables as $n\in\{3,5,8,10\}$. The total number of peaks in MPB and MPBG is $m=5$. In our experiments, we consider both small and large changes by setting the control parameters as $(\tilde{h}=1.0,\tilde{s}=1.0)$ and $(\tilde{h}=5.0,\tilde{s}=7.0)$ respectively.

    \item\underline{Computational budget}: In our experiments, we consider $T=10$ time steps in total and each time step is allocated with an extremely limited FEs. Specifically, $N_\mathrm{FE}=2\times(11n-1)$ at the first time step with $11n-1$ for initial sampling. As for the follow up time steps, $N_\mathrm{FE}=9n$ of which $2n$ FEs are used for the initialization purpose.

    \item\underline{Parameters of peer algorithms}: \textcolor{red}{For \our, the number of clusters $k=3$, the scale of the local search $\kappa=5$ and the closeness threshold $\epsilon_d=0.01$. For \texttt{DIN}, $s^2=2.0$. For \texttt{MBO}, $c=1.0$. For \texttt{TV-GP-UCB}, $\epsilon=1/\sqrt{5}$.}
\end{itemize}


\section{Experimental Results}
\label{sec:experiments}

Our empirical study conducted in this section aims to answer the following research questions (RQs).
\begin{itemize}
    \item\textit{\underline{RQ1}}: How is the performance of \our\ compared against the other peer algorithms at different dimensions?
    \item\textit{\underline{RQ2}}: What is the effect brought by the significance of the change of environments?
    \item\textit{\underline{RQ3}}: What is the benefit of hierarchical version of MOGP versus the conventional LMC?
    \item\textit{\underline{RQ4}}: What is the benefit of the adaptive source data selection?
    \item\textit{\underline{RQ5}}: What is the benefit of the warm starting initialization mechanism?
    \item\textit{\underline{RQ6}}: What is the benefit of the hybrid DE with local search for the optimization of acquisition function?
    \item\textit{\underline{RQ7}}: How is the performance of \our\ on solving a real-world problem in dynamic environments?
\end{itemize}

\subsection{Performance Comparisons with Peer Algorithms}
\label{sec:rq12}

In this subsection, we investigate the performance comparison of our proposed \our\ against the other peer algorithms.

\subsubsection{Methods}
\label{sec:methods_rq12}

To address RQ1 and RQ2, the results are presented from the following five aspects.
\begin{itemize}
    \item First, the statistical comparison results on $\bar{\epsilon}_\mathrm{f}$ and $\bar{\epsilon}_\mathrm{t}$ are given in Tables~\ref{tab:mpb_rq1} and~\ref{tab:mpbg_rq1}.

    \item Then, we apply the $A_{12}$ effect size to further understand the performance difference between \our\ and the other peer algorithms on $\bar{\epsilon}_\mathrm{f}$ and $\bar{\epsilon}_\mathrm{t}$. Given the large amount of comparisons, we pull $9\times 16$ results collected from $A_{12}$ effect size together and calculate the percentage of different effect sizes obtained by a pair of dueling algorithms. These are presented as bar charts in~\pref{fig:a12_rq12}.

    \item To have an overall comparison across different benchmark test problems, we apply the Scott-Knott test to rank the performance of different algorithms. To facilitate a better interpretation of many comparison results, we pull $9\times 16$ comparison results collected from the Scott-Knott test together along with their variances as the bar charts with error bars in~\pref{fig:sk_rq12}.

    \item To facilitate the understanding of the convergence rates of different algorithms during the dynamic optimization process, we develop two dedicated metrics as follows.
        \begin{itemize}
            \item First, we keep a record of the loss w.r.t. the corresponding optimum at a given FE as:
                \begin{equation}
                    \mathcal{L}(\mathbf{x},t)=f(\mathbf{x}^\ast,t)-f(\bar{\mathbf{x}}^\ast,t),
                \end{equation}
                where $(\bar{\mathbf{x}}^\ast,t)$ is the best solution found at the corresponding FE of the $t$-th time step. 
            \item The other metric evaluates the difference of the budget required by a peer algorithm w.r.t. the best algorithm at each time step. It is calculated as:
                \begin{equation}
                    \rho_\mathrm{c}=\frac{1}{T}\sum_{t=1}^T \frac{N_\mathrm{FE}^t}{\tilde{N}_{\mathrm{FE}}^{t\ast}},
                \end{equation}
                where $\tilde{N}_\mathrm{FE}^{t\ast}$ is the number of FEs consumed by the best algorithm at the $t$-th time step to obtain its best solution $(\tilde{\mathbf{x}}^\ast,t)$. $N^t_\mathrm{FE}$ is the number of FEs required by one of the other peer algorithms to achieve $f(\tilde{\mathbf{x}}^\ast,t)$. Note that $N^t_{FE}$ can be up to $8\times N_\mathrm{FE}$ in our experiments no matter whether the corresponding peer algorithm obtains $f(\tilde{\mathbf{x}}^\ast,t)$ or not. In practice, $\rho_\mathrm{c}\geq 1$ and the larger $\rho_\mathrm{c}$ is, the worse the corresponding peer algorithm is.
        \end{itemize}

    \item Last but not the least, we develop another metric $\rho_\mathrm{t}$ to investigate the effectiveness brought by the knowledge transfer, i.e., leveraging historical data from the previous optimization practice. It is calculated as:
        \begin{equation}
            \rho_\mathrm{t}=\frac{1}{T}\sum_{t=1}^T\frac{N_{\mathrm{FE}}^t}{N_\mathrm{c}^t},
        \end{equation}
        where $N_\mathrm{c}^t$ is the number of FEs required by \texttt{RBO}, which completely ignores the historical data but restarts itself from scratch after each change, to obtain its best solution $(\mathbf{x}_\mathrm{c}^\ast,t)$ at the $t$-th time step. $N_{\mathrm{FE}}^t$ is the number of FEs required by one of the other peer algorithms that leverages historical information to a certain extent to achieve $f(\mathbf{x}_\mathrm{c}^\ast,t)$. In practice, the corresponding knowledge transfer approach is regarded as effective when $0<\rho_\mathrm{t}<1$ (the smaller $\rho_\mathrm{t}$ is, the better knowledge transfer is); otherwise it is negative to the underlying baseline optimization routine~\cite{SunL20,NieGL20,YangHL21}.
\end{itemize}

\subsubsection{Results}
\label{sec:results_rq12}

Let us interpret the results according to the methods introduced in~\pref{sec:methods_rq12}.
\begin{itemize}
    \item From the comparison results on $\bar{\epsilon}_\mathrm{f}$ and $\bar{\epsilon}_\mathrm{t}$ shown in Tables~\ref{tab:mpb_rq1} and~\ref{tab:mpbg_rq1}, we can see that our proposed \our\ is always the best algorithm in all comparisons without any exception. Note that all the better results are of statistical significance according to the Wilcoxon signed-rank test.

        \begin{table*}[htbp]
            \caption{Performance comparison results of $\overline{\epsilon}_f$ and $\overline{\epsilon}_t$ between \our\ and the other $9$ peer algorithms on the MPB problems}
            \label{tab:mpb_rq1}
            \resizebox{1.0\textwidth}{!}{ 
            \begin{tabular}{c|cc|cc|cc|cc}
                \hline
                \multirow{2}{*}{$\epsilon_t$} & \multicolumn{2}{c|}{$n=3$}                                                         & \multicolumn{2}{c|}{$n=5$}                                                         & \multicolumn{2}{c|}{$n=8$}                                                         & \multicolumn{2}{c}{$n=10$}                                                        \\ \cline{2-9} 
                & \multicolumn{1}{c|}{$\tilde{h}=1.0,\tilde{s}=1.0$} & $\tilde{h}=5.0,\tilde{s}=7.0$ & \multicolumn{1}{c|}{$\tilde{h}=1.0,\tilde{s}=1.0$} & $\tilde{h}=5.0,\tilde{s}=7.0$ & \multicolumn{1}{c|}{$\tilde{h}=1.0,\tilde{s}=1.0$} & $\tilde{h}=5.0,\tilde{s}=7.0$ & \multicolumn{1}{c|}{$\tilde{h}=1.0,\tilde{s}=1.0$} & $\tilde{h}=5.0,\tilde{s}=7.0$ \\ \hline
                \texttt{TBO-II}               & \multicolumn{1}{c|}{\bb{1.395E+1(6.51E+0)}}       & \bb{2.000E+1(3.73E+0)}       & \multicolumn{1}{c|}{\bb{1.561E+1(1.22E+1)}}       & \bb{2.628E+1(6.52E+0)}       & \multicolumn{1}{c|}{\bb{1.867E+1(9.86E+0)}}       & \bb{2.476E+1(5.32E+0)}       & \multicolumn{1}{c|}{\bb{1.863E+1(1.10E+1)}}       & \bb{2.211E+1(5.21E+0)}       \\ \hline
                \texttt{RBO}                  & \multicolumn{1}{c|}{3.774E+1(2.00E+1)$^\dag$}     & 3.894E+1(1.28E+1)$^\dag$     & \multicolumn{1}{c|}{4.028E+1(1.95E+1)$^\dag$}     & 4.460E+1(1.03E+1)$^\dag$     & \multicolumn{1}{c|}{5.160E+1(2.84E+1)$^\dag$}     & 5.463E+1(1.47E+1)$^\dag$     & \multicolumn{1}{c|}{5.372E+1(1.98E+1)$^\dag$}     & 7.219E+1(1.74E+1)$^\dag$     \\ \hline
                \texttt{CBO}                  & \multicolumn{1}{c|}{3.636E+1(1.32E+1)$^\dag$}     & 4.188E+1(1.45E+1)$^\dag$     & \multicolumn{1}{c|}{6.195E+1(1.90E+1)$^\dag$}     & 8.597E+1(2.41E+1)$^\dag$     & \multicolumn{1}{c|}{8.607E+1(2.90E+1)$^\dag$}     & 1.211E+2(2.44E+1)$^\dag$     & \multicolumn{1}{c|}{8.539E+1(3.23E+1)$^\dag$}     & 1.568E+2(5.36E+1)$^\dag$     \\ \hline
                \texttt{DIN}                  & \multicolumn{1}{c|}{2.899E+1(1.05E+1)$^\dag$}     & 3.996E+1(1.23E+1)$^\dag$     & \multicolumn{1}{c|}{5.073E+1(2.74E+1)$^\dag$}     & 6.136E+1(2.06E+1)$^\dag$     & \multicolumn{1}{c|}{6.329E+1(2.75E+1)$^\dag$}     & 9.385E+1(3.44E+1)$^\dag$     & \multicolumn{1}{c|}{6.010E+1(1.68E+1)$^\dag$}     & 9.607E+1(2.07E+1)$^\dag$     \\ \hline
                \texttt{MBO}                  & \multicolumn{1}{c|}{2.990E+1(1.74E+1)$^\dag$}     & 3.562E+1(1.60E+1)$^\dag$     & \multicolumn{1}{c|}{3.449E+1(1.76E+1)$^\dag$}     & 5.887E+1(1.27E+1)$^\dag$     & \multicolumn{1}{c|}{5.012E+1(6.63E+0)$^\dag$}     & 9.617E+1(3.82E+1)$^\dag$     & \multicolumn{1}{c|}{4.889E+1(1.43E+1)$^\dag$}     & 1.144E+2(4.58E+1)$^\dag$     \\ \hline
                \texttt{TV-GP-UCB}            & \multicolumn{1}{c|}{2.598E+1(9.58E+0)$^\dag$}     & 3.059E+1(8.46E+0)$^\dag$     & \multicolumn{1}{c|}{3.105E+1(1.92E+1)$^\dag$}     & 4.410E+1(1.41E+1)$^\dag$     & \multicolumn{1}{c|}{3.839E+1(1.73E+1)$^\dag$}     & 6.954E+1(3.12E+1)$^\dag$     & \multicolumn{1}{c|}{4.362E+1(1.81E+1)$^\dag$}     & 7.524E+1(3.11E+1)$^\dag$     \\ \hline
                \texttt{Box+GP}               & \multicolumn{1}{c|}{5.639E+1(1.49E+1)$^\dag$}     & 6.901E+1(1.90E+1)$^\dag$     & \multicolumn{1}{c|}{7.752E+1(2.44E+1)$^\dag$}     & 1.105E+2(3.48E+1)$^\dag$     & \multicolumn{1}{c|}{1.037E+2(3.32E+1)$^\dag$}     & 1.422E+2(3.98E+1)$^\dag$     & \multicolumn{1}{c|}{1.055E+2(3.84E+1)$^\dag$}     & 1.677E+2(4.70E+1)$^\dag$     \\ \hline
                \texttt{RGPE}                 & \multicolumn{1}{c|}{2.041E+1(1.17E+1)$^\dag$}     & 2.125E+1(7.21E+0)$^\dag$     & \multicolumn{1}{c|}{2.580E+1(8.92E+0)$^\dag$}     & 3.012E+1(1.35E+1)$^\dag$     & \multicolumn{1}{c|}{2.870E+1(1.54E+1)$^\dag$}     & 3.570E+1(1.29E+1)$^\dag$     & \multicolumn{1}{c|}{2.806E+1(1.58E+1)$^\dag$}     & 4.271E+1(8.63E+0)$^\dag$     \\ \hline
                \texttt{BHGP}                 & \multicolumn{1}{c|}{2.480E+1(1.41E+1)$^\dag$}     & 3.340E+1(1.20E+1)$^\dag$     & \multicolumn{1}{c|}{3.076E+1(1.81E+1)$^\dag$}     & 4.591E+1(1.22E+1)$^\dag$     & \multicolumn{1}{c|}{4.407E+1(1.88E+1)$^\dag$}     & 6.596E+1(3.26E+1)$^\dag$     & \multicolumn{1}{c|}{4.530E+1(1.52E+1)$^\dag$}     & 7.310E+1(1.62E+1)$^\dag$     \\ \hline
                \texttt{TBO}                  & \multicolumn{1}{c|}{1.892E+1(8.17E+0)$^\dag$}     & 2.370E+1(1.02E+1)$^\dag$     & \multicolumn{1}{c|}{2.356E+1(1.57E+1)$^\dag$}     & 3.322E+1(1.01E+1)$^\dag$     & \multicolumn{1}{c|}{3.324E+1(1.57E+1)$^\dag$}     & 4.491E+1(1.97E+1)$^\dag$     & \multicolumn{1}{c|}{3.041E+1(8.52E+0)$^\dag$}     & 4.754E+1(2.23E+1)$^\dag$     \\ \hline\hline
                \multirow{2}{*}{$\epsilon_f$} & \multicolumn{2}{c|}{$n=3$}                                                         & \multicolumn{2}{c|}{$n=5$}                                                         & \multicolumn{2}{c|}{$n=8$}                                                         & \multicolumn{2}{c}{$n=10$}                                                        \\ \cline{2-9} 
                & \multicolumn{1}{c|}{$\tilde{h}=1.0,\tilde{s}=1.0$} & $\tilde{h}=5.0,\tilde{s}=7.0$ & \multicolumn{1}{c|}{$\tilde{h}=1.0,\tilde{s}=1.0$} & $\tilde{h}=5.0,\tilde{s}=7.0$ & \multicolumn{1}{c|}{$\tilde{h}=1.0,\tilde{s}=1.0$} & $\tilde{h}=5.0,\tilde{s}=7.0$ & \multicolumn{1}{c|}{$\tilde{h}=1.0,\tilde{s}=1.0$} & $\tilde{h}=5.0,\tilde{s}=7.0$ \\ \hline
                \texttt{TBO-II}               & \multicolumn{1}{c|}{\bb{3.690E+1(1.59E+1)}}       & \bb{4.191E+1(1.09E+1)}       & \multicolumn{1}{c|}{\bb{5.257E+1(1.80E+1)}}       & \bb{5.669E+1(1.50E+1)}       & \multicolumn{1}{c|}{\bb{6.254E+1(2.60E+1)}}       & \bb{6.918E+1(1.21E+1)}       & \multicolumn{1}{c|}{\bb{6.906E+1(2.30E+1)}}       & \bb{7.647E+1(2.03E+1)}       \\ \hline
                \texttt{RBO}                  & \multicolumn{1}{c|}{5.806E+1(1.65E+1)$^\dag$}     & 6.437E+1(2.10E+1)$^\dag$     & \multicolumn{1}{c|}{7.656E+1(3.48E+1)$^\dag$}     & 9.685E+1(3.05E+1)$^\dag$     & \multicolumn{1}{c|}{1.087E+2(5.82E+1)$^\dag$}     & 1.318E+2(3.32E+1)$^\dag$     & \multicolumn{1}{c|}{1.379E+2(4.30E+1)$^\dag$}     & 1.607E+2(2.31E+1)$^\dag$     \\ \hline
                \texttt{CBO}                  & \multicolumn{1}{c|}{5.426E+1(1.92E+1)$^\dag$}     & 6.117E+1(1.07E+1)$^\dag$     & \multicolumn{1}{c|}{7.721E+1(3.17E+1)$^\dag$}     & 9.705E+1(2.21E+1)$^\dag$     & \multicolumn{1}{c|}{9.973E+1(4.16E+1)$^\dag$}     & 1.332E+2(3.22E+1)$^\dag$     & \multicolumn{1}{c|}{1.129E+2(3.62E+1)$^\dag$}     & 1.657E+2(4.01E+1)$^\dag$     \\ \hline
                \texttt{DIN}                  & \multicolumn{1}{c|}{5.065E+1(1.97E+1)$^\dag$}     & 5.202E+1(1.69E+1)$^\dag$     & \multicolumn{1}{c|}{6.876E+1(3.03E+1)$^\dag$}     & 8.319E+1(1.93E+1)$^\dag$     & \multicolumn{1}{c|}{8.579E+1(4.21E+1)$^\dag$}     & 1.137E+2(2.86E+1)$^\dag$     & \multicolumn{1}{c|}{9.184E+1(3.16E+1)$^\dag$}     & 1.298E+2(2.74E+1)$^\dag$     \\ \hline
                \texttt{MBO}                  & \multicolumn{1}{c|}{4.897E+1(1.91E+1)$^\dag$}     & 5.529E+1(1.39E+1)$^\dag$     & \multicolumn{1}{c|}{6.288E+1(2.47E+1)$^\dag$}     & 8.921E+1(2.16E+1)$^\dag$     & \multicolumn{1}{c|}{7.965E+1(3.96E+1)$^\dag$}     & 1.235E+2(3.84E+1)$^\dag$     & \multicolumn{1}{c|}{9.696E+1(2.48E+1)$^\dag$}     & 1.488E+2(3.98E+1)$^\dag$     \\ \hline
                \texttt{TV-GP-UCB}            & \multicolumn{1}{c|}{4.451E+1(1.19E+1)$^\dag$}     & 4.739E+1(1.42E+1)$^\dag$     & \multicolumn{1}{c|}{6.311E+1(2.59E+1)$^\dag$}     & 8.507E+1(2.35E+1)$^\dag$     & \multicolumn{1}{c|}{7.781E+1(3.31E+1)$^\dag$}     & 1.104E+2(3.40E+1)$^\dag$     & \multicolumn{1}{c|}{9.262E+1(3.16E+1)$^\dag$}     & 1.218E+2(1.98E+1)$^\dag$     \\ \hline
                \texttt{Box+GP}               & \multicolumn{1}{c|}{5.877E+1(1.74E+1)$^\dag$}     & 6.794E+1(2.04E+1)$^\dag$     & \multicolumn{1}{c|}{7.743E+1(3.04E+1)$^\dag$}     & 1.052E+2(3.08E+1)$^\dag$     & \multicolumn{1}{c|}{1.066E+2(5.20E+1)$^\dag$}     & 1.321E+2(3.10E+1)$^\dag$     & \multicolumn{1}{c|}{1.169E+2(3.67E+1)$^\dag$}     & 1.606E+2(4.08E+1)$^\dag$     \\ \hline
                \texttt{RGPE}                 & \multicolumn{1}{c|}{4.032E+1(1.02E+1)$^\dag$}     & 4.503E+1(9.05E+0)$^\dag$     & \multicolumn{1}{c|}{5.412E+1(1.85E+1)$^\dag$}     & 6.937E+1(1.28E+1)$^\dag$     & \multicolumn{1}{c|}{6.950E+1(3.79E+1)$^\dag$}     & 8.776E+1(2.88E+1)$^\dag$     & \multicolumn{1}{c|}{7.709E+1(2.70E+1)$^\dag$}     & 1.065E+2(1.37E+1)$^\dag$     \\ \hline
                \texttt{BHGP}                 & \multicolumn{1}{c|}{4.427E+1(1.16E+1)$^\dag$}     & 4.885E+1(1.13E+1)$^\dag$     & \multicolumn{1}{c|}{5.512E+1(2.23E+1)$^\dag$}     & 7.797E+1(1.97E+1)$^\dag$     & \multicolumn{1}{c|}{7.305E+1(3.44E+1)$^\dag$}     & 1.030E+2(3.02E+1)$^\dag$     & \multicolumn{1}{c|}{8.009E+1(2.54E+1)$^\dag$}     & 1.126E+2(3.51E+1)$^\dag$     \\ \hline
                \texttt{TBO}                  & \multicolumn{1}{c|}{4.459E+1(1.33E+1)$^\dag$}     & 4.680E+1(1.29E+1)$^\dag$     & \multicolumn{1}{c|}{5.499E+1(1.98E+1)$^\dag$}     & 6.945E+1(1.33E+1)$^\dag$     & \multicolumn{1}{c|}{7.075E+1(3.52E+1)$^\dag$}     & 9.569E+1(2.97E+1)$^\dag$     & \multicolumn{1}{c|}{7.468E+1(2.32E+1)$^\dag$}     & 1.100E+2(2.48E+1)$^\dag$     \\ \hline
            \end{tabular}
            }
            \begin{tablenotes}
                \scriptsize
            \item $^\dagger$ indicates that \our\ is significantly better than the corresponding peer algorithm according to the Wilcoxon signed-rank test at the 5\% significance level.
            \end{tablenotes}
        \end{table*}

        \begin{table*}[htbp]
            \caption{Performance comparison results of $\overline{\epsilon}_f$ and $\overline{\epsilon}_t$ between \our\ and the other $9$ peer algorithms on the MPBG problems}
            \label{tab:mpbg_rq1}
            \resizebox{1.0\textwidth}{!}{
            \begin{tabular}{c|cc|cc|cc|cc}
                \hline
                \multirow{2}{*}{$\epsilon_t$} & \multicolumn{2}{c|}{$n=3$}                                                         & \multicolumn{2}{c|}{$n=5$}                                                         & \multicolumn{2}{c|}{$n=8$}                                                         & \multicolumn{2}{c}{$n=10$}                                                        \\ \cline{2-9} 
                & \multicolumn{1}{c|}{$\tilde{h}=1.0,\tilde{s}=1.0$} & $\tilde{h}=5.0,\tilde{s}=7.0$ & \multicolumn{1}{c|}{$\tilde{h}=1.0,\tilde{s}=1.0$} & $\tilde{h}=5.0,\tilde{s}=7.0$ & \multicolumn{1}{c|}{$\tilde{h}=1.0,\tilde{s}=1.0$} & $\tilde{h}=5.0,\tilde{s}=7.0$ & \multicolumn{1}{c|}{$\tilde{h}=1.0,\tilde{s}=1.0$} & $\tilde{h}=5.0,\tilde{s}=7.0$ \\ \hline
                \texttt{DETO}                 & \multicolumn{1}{c|}{\bb{1.229E+1(1.27E+1)}}       & \bb{1.328E+1(1.82E+1)}       & \multicolumn{1}{c|}{\bb{1.306E+1(1.98E+1)}}       & \bb{2.286E+1(2.67E+1)}       & \multicolumn{1}{c|}{\bb{1.569E+1(1.80E+1)}}       & \bb{2.217E+1(1.97E+1)}       & \multicolumn{1}{c|}{\bb{1.472E+1(2.26E+1)}}       & \bb{2.042E+1(2.00E+1)}       \\ \hline
                \texttt{RBO}                  & \multicolumn{1}{c|}{2.626E+1(3.73E+1)$^\dag$}     & 3.868E+1(3.93E+1)$^\dag$     & \multicolumn{1}{c|}{4.137E+1(5.89E+1)$^\dag$}     & 4.327E+1(4.11E+1)$^\dag$     & \multicolumn{1}{c|}{4.391E+1(4.72E+1)$^\dag$}     & 4.718E+1(4.67E+1)$^\dag$     & \multicolumn{1}{c|}{4.085E+1(4.49E+1)$^\dag$}     & 5.869E+1(6.02E+1)$^\dag$     \\ \hline
                \texttt{CBO}                  & \multicolumn{1}{c|}{3.430E+1(3.91E+1)$^\dag$}     & 3.849E+1(3.77E+1)$^\dag$     & \multicolumn{1}{c|}{5.795E+1(6.15E+1)$^\dag$}     & 6.809E+1(7.42E+1)$^\dag$     & \multicolumn{1}{c|}{7.899E+1(7.60E+1)$^\dag$}     & 1.089E+2(1.05E+2)$^\dag$     & \multicolumn{1}{c|}{6.509E+1(7.33E+1)$^\dag$}     & 1.155E+2(1.49E+2)$^\dag$     \\ \hline
                \texttt{DIN}                  & \multicolumn{1}{c|}{2.653E+1(2.50E+1)$^\dag$}     & 3.923E+1(3.66E+1)$^\dag$     & \multicolumn{1}{c|}{4.478E+1(5.39E+1)$^\dag$}     & 5.646E+1(4.99E+1)$^\dag$     & \multicolumn{1}{c|}{4.929E+1(5.46E+1)$^\dag$}     & 7.399E+1(8.28E+1)$^\dag$     & \multicolumn{1}{c|}{5.117E+1(5.06E+1)$^\dag$}     & 8.622E+1(8.05E+1)$^\dag$     \\ \hline
                \texttt{MBO}                  & \multicolumn{1}{c|}{2.287E+1(2.89E+1)$^\dag$}     & 3.464E+1(3.48E+1)$^\dag$     & \multicolumn{1}{c|}{3.168E+1(3.95E+1)$^\dag$}     & 5.077E+1(4.95E+1)$^\dag$     & \multicolumn{1}{c|}{4.328E+1(4.65E+1)$^\dag$}     & 7.923E+1(9.40E+1)$^\dag$     & \multicolumn{1}{c|}{4.070E+1(3.75E+1)$^\dag$}     & 8.891E+1(1.01E+2)$^\dag$     \\ \hline
                \texttt{TV-GP-UCB}            & \multicolumn{1}{c|}{2.122E+1(2.48E+1)$^\dag$}     & 2.399E+1(2.74E+1)$^\dag$     & \multicolumn{1}{c|}{3.105E+1(3.81E+1)$^\dag$}     & 3.980E+1(4.04E+1)$^\dag$     & \multicolumn{1}{c|}{3.474E+1(3.85E+1)$^\dag$}     & 5.885E+1(7.33E+1)$^\dag$     & \multicolumn{1}{c|}{3.401E+1(4.15E+1)$^\dag$}     & 5.715E+1(6.56E+1)$^\dag$     \\ \hline
                \texttt{Box+GP}               & \multicolumn{1}{c|}{4.903E+1(5.13E+1)$^\dag$}     & 6.718E+1(5.90E+1)$^\dag$     & \multicolumn{1}{c|}{7.694E+1(8.72E+1)$^\dag$}     & 8.971E+1(9.42E+1)$^\dag$     & \multicolumn{1}{c|}{7.415E+1(9.97E+1)$^\dag$}     & 1.272E+2(1.12E+2)$^\dag$     & \multicolumn{1}{c|}{8.440E+1(7.96E+1)$^\dag$}     & 1.380E+2(1.45E+2)$^\dag$     \\ \hline
                \texttt{RGPE}                 & \multicolumn{1}{c|}{1.301E+1(1.95E+1)$^\dag$}     & 2.115E+1(2.38E+1)$^\dag$     & \multicolumn{1}{c|}{2.509E+1(2.34E+1)$^\dag$}     & 2.568E+1(2.48E+1)$^\dag$     & \multicolumn{1}{c|}{2.212E+1(2.40E+1)$^\dag$}     & 3.191E+1(2.97E+1)$^\dag$     & \multicolumn{1}{c|}{2.105E+1(2.51E+1)$^\dag$}     & 3.948E+1(3.55E+1)$^\dag$     \\ \hline
                \texttt{BHGP}                 & \multicolumn{1}{c|}{2.281E+1(2.46E+1)$^\dag$}     & 2.467E+1(3.23E+1)$^\dag$     & \multicolumn{1}{c|}{2.989E+1(4.04E+1)$^\dag$}     & 3.933E+1(4.08E+1)$^\dag$     & \multicolumn{1}{c|}{3.950E+1(4.64E+1)$^\dag$}     & 6.312E+1(6.84E+1)$^\dag$     & \multicolumn{1}{c|}{3.408E+1(3.99E+1)$^\dag$}     & 7.053E+1(6.06E+1)$^\dag$     \\ \hline
                \texttt{TBO}                  & \multicolumn{1}{c|}{1.646E+1(2.10E+1)$^\dag$}     & 2.102E+1(2.62E+1)$^\dag$     & \multicolumn{1}{c|}{2.269E+1(2.98E+1)$^\dag$}     & 2.632E+1(3.14E+1)$^\dag$     & \multicolumn{1}{c|}{2.411E+1(2.98E+1)$^\dag$}     & 3.622E+1(4.18E+1)$^\dag$     & \multicolumn{1}{c|}{2.603E+1(2.70E+1)$^\dag$}     & 3.822E+1(4.19E+1)$^\dag$     \\ \hline\hline
                \multirow{2}{*}{$\epsilon_f$} & \multicolumn{2}{c|}{$n=3$}                                                         & \multicolumn{2}{c|}{$n=5$}                                                         & \multicolumn{2}{c|}{$n=8$}                                                         & \multicolumn{2}{c}{$n=10$}                                                        \\ \cline{2-9} 
                & \multicolumn{1}{c|}{$\tilde{h}=1.0,\tilde{s}=1.0$} & $\tilde{h}=5.0,\tilde{s}=7.0$ & \multicolumn{1}{c|}{$\tilde{h}=1.0,\tilde{s}=1.0$} & $\tilde{h}=5.0,\tilde{s}=7.0$ & \multicolumn{1}{c|}{$\tilde{h}=1.0,\tilde{s}=1.0$} & $\tilde{h}=5.0,\tilde{s}=7.0$ & \multicolumn{1}{c|}{$\tilde{h}=1.0,\tilde{s}=1.0$} & $\tilde{h}=5.0,\tilde{s}=7.0$ \\ \hline
                \texttt{DETO}                 & \multicolumn{1}{c|}{\bb{2.933E+1(3.42E+1)}}       & \bb{3.911E+1(3.97E+1)}       & \multicolumn{1}{c|}{\bb{4.859E+1(5.30E+1)}}       & \bb{5.281E+1(5.22E+1)}       & \multicolumn{1}{c|}{\bb{5.152E+1(6.43E+1)}}       & \bb{6.505E+1(5.89E+1)}       & \multicolumn{1}{c|}{\bb{5.161E+1(6.12E+1)}}       & \bb{6.830E+1(6.74E+1)}       \\ \hline
                \texttt{RBO}                  & \multicolumn{1}{c|}{5.237E+1(5.67E+1)$^\dag$}     & 6.437E+1(6.41E+1)$^\dag$     & \multicolumn{1}{c|}{7.174E+1(9.41E+1)$^\dag$}     & 8.154E+1(8.67E+1)$^\dag$     & \multicolumn{1}{c|}{9.755E+1(1.23E+2)$^\dag$}     & 1.141E+2(1.12E+2)$^\dag$     & \multicolumn{1}{c|}{9.258E+1(1.22E+2)$^\dag$}     & 1.470E+2(1.35E+2)$^\dag$     \\ \hline
                \texttt{CBO}                  & \multicolumn{1}{c|}{4.270E+1(4.87E+1)$^\dag$}     & 5.939E+1(5.05E+1)$^\dag$     & \multicolumn{1}{c|}{7.384E+1(9.02E+1)$^\dag$}     & 8.334E+1(7.84E+1)$^\dag$     & \multicolumn{1}{c|}{8.967E+1(1.03E+2)$^\dag$}     & 1.182E+2(1.11E+2)$^\dag$     & \multicolumn{1}{c|}{8.410E+1(1.05E+2)$^\dag$}     & 1.389E+2(1.45E+2)$^\dag$     \\ \hline
                \texttt{DIN}                  & \multicolumn{1}{c|}{3.856E+1(4.84E+1)$^\dag$}     & 5.123E+1(5.21E+1)$^\dag$     & \multicolumn{1}{c|}{6.186E+1(8.30E+1)$^\dag$}     & 7.142E+1(6.81E+1)$^\dag$     & \multicolumn{1}{c|}{7.378E+1(8.52E+1)$^\dag$}     & 9.997E+1(9.76E+1)$^\dag$     & \multicolumn{1}{c|}{7.496E+1(8.31E+1)$^\dag$}     & 1.187E+2(1.05E+2)$^\dag$     \\ \hline
                \texttt{MBO}                  & \multicolumn{1}{c|}{3.826E+1(4.39E+1)$^\dag$}     & 5.282E+1(4.87E+1)$^\dag$     & \multicolumn{1}{c|}{6.069E+1(7.79E+1)$^\dag$}     & 7.410E+1(7.02E+1)$^\dag$     & \multicolumn{1}{c|}{7.101E+1(8.07E+1)$^\dag$}     & 1.133E+2(1.10E+2)$^\dag$     & \multicolumn{1}{c|}{7.135E+1(8.00E+1)$^\dag$}     & 1.296E+2(1.28E+2)$^\dag$     \\ \hline
                \texttt{TV-GP-UCB}            & \multicolumn{1}{c|}{3.989E+1(4.17E+1)$^\dag$}     & 4.263E+1(4.50E+1)$^\dag$     & \multicolumn{1}{c|}{5.938E+1(7.07E+1)$^\dag$}     & 7.374E+1(7.15E+1)$^\dag$     & \multicolumn{1}{c|}{7.387E+1(7.75E+1)$^\dag$}     & 1.096E+2(1.12E+2)$^\dag$     & \multicolumn{1}{c|}{7.082E+1(8.68E+1)$^\dag$}     & 1.144E+2(9.82E+1)$^\dag$     \\ \hline
                \texttt{Box+GP}               & \multicolumn{1}{c|}{5.116E+1(5.03E+1)$^\dag$}     & 6.404E+1(6.17E+1)$^\dag$     & \multicolumn{1}{c|}{7.703E+1(9.24E+1)$^\dag$}     & 8.417E+1(9.01E+1)$^\dag$     & \multicolumn{1}{c|}{8.236E+1(1.04E+2)$^\dag$}     & 1.269E+2(1.07E+2)$^\dag$     & \multicolumn{1}{c|}{8.490E+1(9.83E+1)$^\dag$}     & 1.322E+2(1.34E+2)$^\dag$     \\ \hline
                \texttt{RGPE}                 & \multicolumn{1}{c|}{3.256E+1(3.36E+1)$^\dag$}     & 4.272E+1(4.08E+1)$^\dag$     & \multicolumn{1}{c|}{5.047E+1(6.82E+1)$^\dag$}     & 5.910E+1(5.85E+1)$^\dag$     & \multicolumn{1}{c|}{6.088E+1(7.68E+1)$^\dag$}     & 7.478E+1(7.77E+1)$^\dag$     & \multicolumn{1}{c|}{6.000E+1(6.64E+1)$^\dag$}     & 9.888E+1(8.66E+1)$^\dag$     \\ \hline
                \texttt{BHGP}                 & \multicolumn{1}{c|}{3.666E+1(3.98E+1)$^\dag$}     & 4.669E+1(4.77E+1)$^\dag$     & \multicolumn{1}{c|}{5.432E+1(6.32E+1)$^\dag$}     & 6.923E+1(6.90E+1)$^\dag$     & \multicolumn{1}{c|}{6.881E+1(7.95E+1)$^\dag$}     & 9.749E+1(1.00E+2)$^\dag$     & \multicolumn{1}{c|}{6.336E+1(7.37E+1)$^\dag$}     & 1.041E+2(9.08E+1)$^\dag$     \\ \hline
                \texttt{TBO}                  & \multicolumn{1}{c|}{3.618E+1(3.76E+1)$^\dag$}     & 4.307E+1(4.57E+1)$^\dag$     & \multicolumn{1}{c|}{5.231E+1(6.55E+1)$^\dag$}     & 6.502E+1(5.41E+1)$^\dag$     & \multicolumn{1}{c|}{5.808E+1(7.59E+1)$^\dag$}     & 8.284E+1(8.41E+1)$^\dag$     & \multicolumn{1}{c|}{5.875E+1(6.07E+1)$^\dag$}     & 9.955E+1(9.96E+1)$^\dag$     \\ \hline
            \end{tabular}
            }
            \begin{tablenotes}
                \scriptsize
            \item $^\dagger$ indicates that \our\ is significantly better than the corresponding peer algorithm according to the Wilcoxon signed-rank test at the 5\% significance level.
            \end{tablenotes}
        \end{table*}

    \item In view of the overwhelmingly superior performance observed above, it is not surprised to see the stacked bar charts of $A_{12}$ shown in~\pref{fig:a12_rq12} demonstrate that all the better comparison results achieved by \our\ are always classified to be large.
        \begin{figure}[t!]
            \centering
            \includegraphics[width=.7\linewidth]{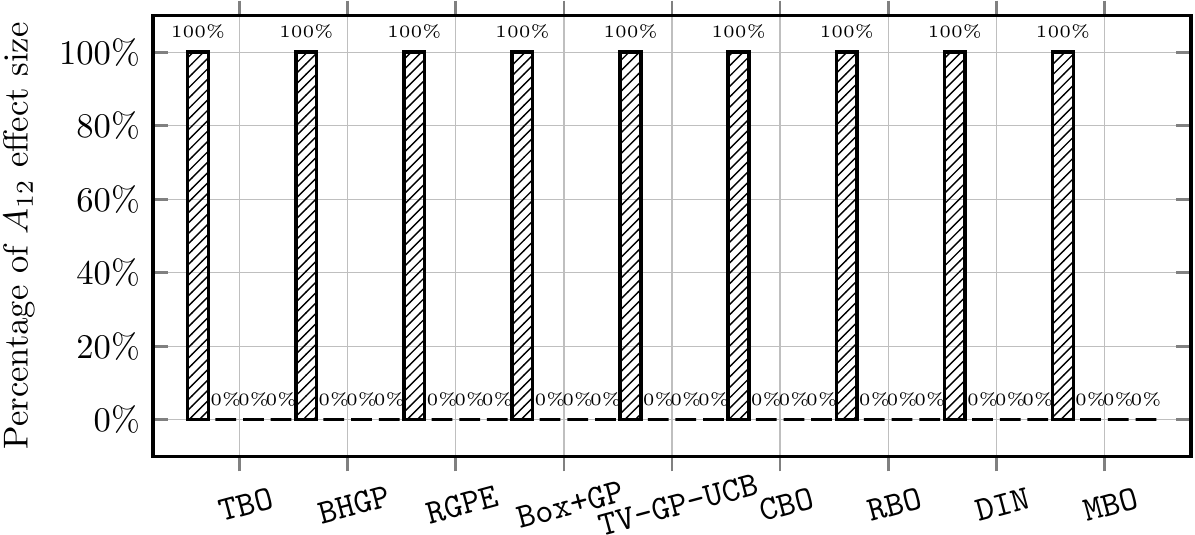}
            \caption{Percentage of the large, medium, small, and equal $A_{12}$ effect size of $\bar{\epsilon}_\mathrm{f}$ and $\bar{\epsilon}_\mathrm{t}$ when comparing \our\ with the other nine peer algorithms.}
            \label{fig:a12_rq12}
        \end{figure}

    \item Likewise, based on the previous two comparison results, it is not difficult to infer the outstanding rank achieved by \our\ according to the Scott-Knott test as shown in Figs.~\ref{fig:sk_rq12} and~\ref{fig:heatmap_rq12}. In addition, we also notice that \texttt{RBO}, \texttt{CBO}, and \texttt{Box+GP} are the worst peer algorithms. As for \texttt{RBO} and \texttt{CBO}, their inferior performance is not surprising as ignoring any changes can hardly enable the BO to adapt to the new problem within a limited amount of FEs. Although \texttt{Box+GP} claimed to use a transfer learning approach to leverage the previous knowledge, its idea of shrinking search space is shown to be a mess when tackling the benchmark problems considered in our experiments. As the predecessor of \our, the performance of \texttt{TBO} is competitive as it is ranked in the second place according to the Scott-Knott test. \texttt{BHGP} and \texttt{RGPE} also achieve a similar performance as \texttt{TBO}.
        \begin{figure}[t!]
            \centering
            \includegraphics[width=.7\linewidth]{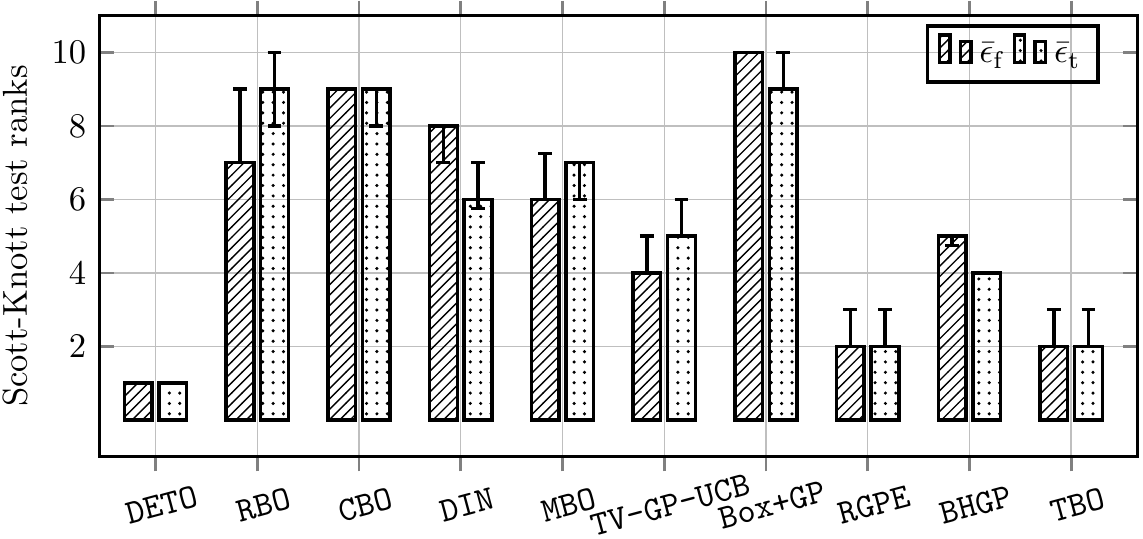}
            \caption{Bar charts with error bars of Scott-Knott test ranks of $\bar{\epsilon}_\mathrm{f}$ and $\bar{\epsilon}_\mathrm{t}$ achieved by each of the ten algorithms (the smaller rank is, the better performance achieves).}
            \label{fig:sk_rq12}
        \end{figure}

        \begin{figure}[t!]
            \centering
            \includegraphics[width=.7\linewidth]{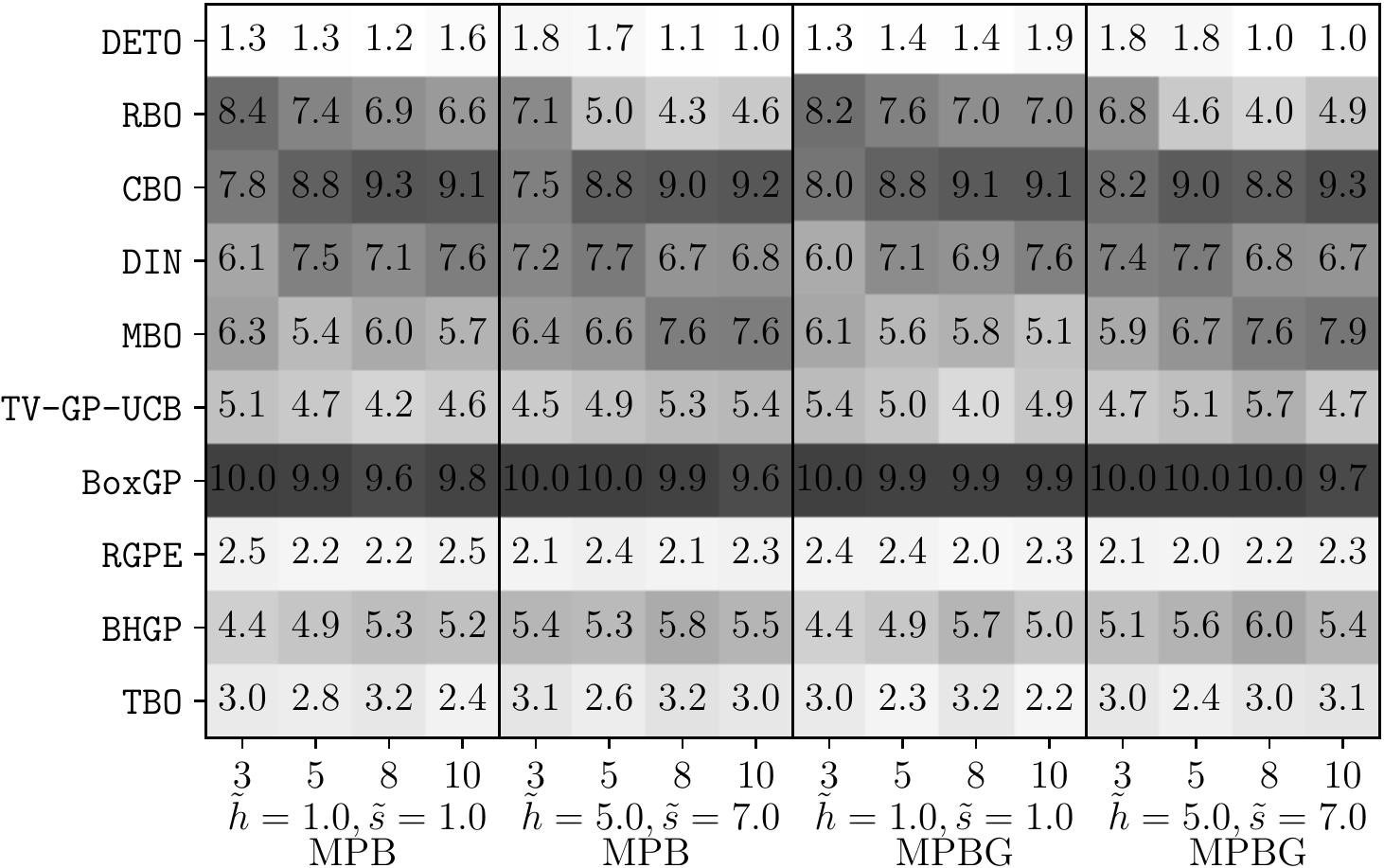}
            \caption{Heatmap of a breakdown of the average Scott-Knott test ranks achieved by ten algorithms (the smaller rank with a lighter cell is, the better performance achieves).}
            \label{fig:heatmap_rq12}
        \end{figure}

    \item From the selected trajectories of $\mathcal{L}(\mathbf{x},t)$ shown in Figs.~\ref{fig:traj_mpb} and~\ref{fig:traj_mpbg} while full results are provided in Figs. 3 and 4 in the Appendix B of the supplemental document, we can see that the performance \our\ is close to the other nine peer algorithms at the first time step. This makes sense because there is no previous data available for \our, whereas we observe a clear jump start since the second time step. This can be attributed to the effectiveness of leveraging \lq knowledge\rq\ collected from the previous optimization practices. Although the other peer algorithms, except \texttt{RBO} and \texttt{CBO}, claimed to have a strategy to transfer knowledge from the historic data, they are not as effective as \our. This might be because these algorithms are not specifically designed for dynamic optimization of which the problem changes overtime. Since only a limited amount of data are collected at each time step, they are not adequate to train a good model~\cite{BillingsleyMLMG20,BillingsleyLMMG20,YuceL21}. 
        \begin{figure}[t!]
            \centering
            \includegraphics[width=.7\linewidth]{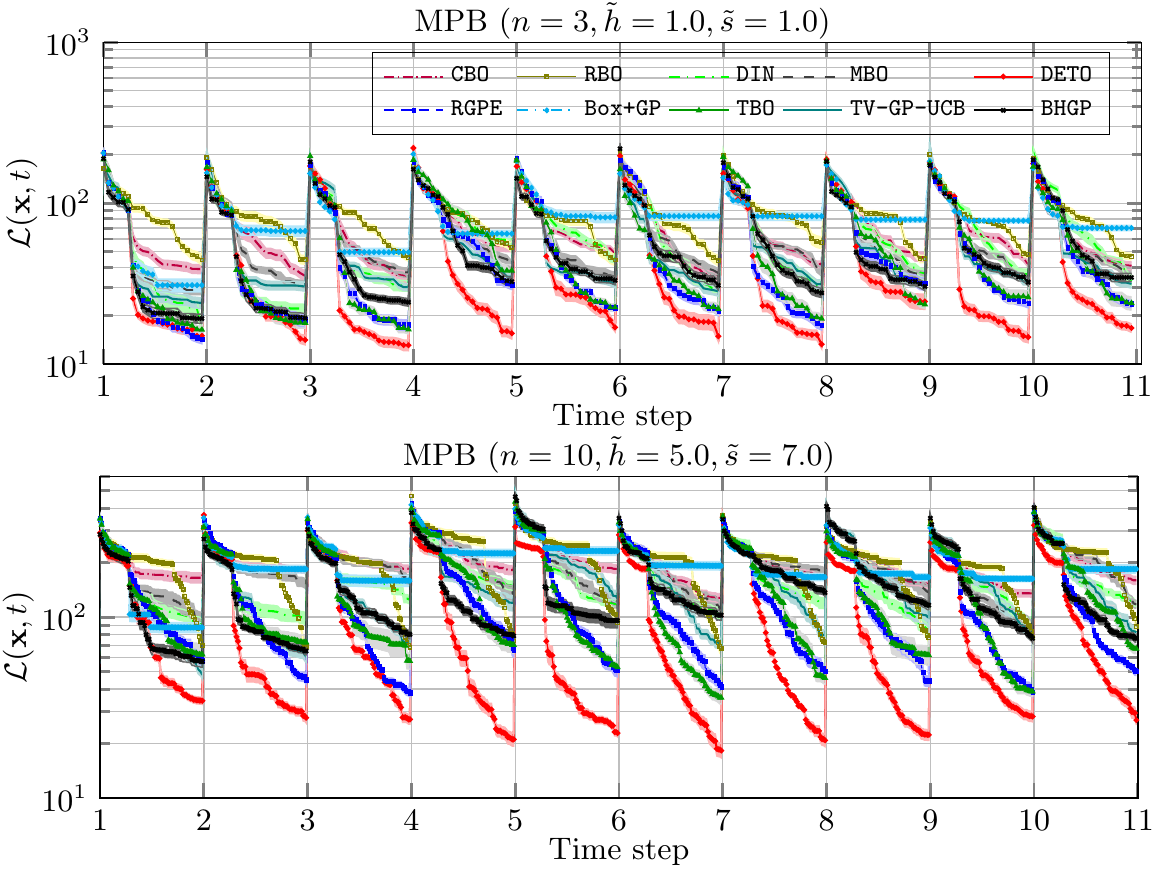}
            \caption{Trajectories of $\mathcal{L}(\mathbf{x},t)$ with confidence bounds over $T=10$ time steps achieved by ten algorithms on two selected MPB instances.}
            \label{fig:traj_mpb}
        \end{figure}

        \begin{figure}[t!]
            \centering
            \includegraphics[width=.7\linewidth]{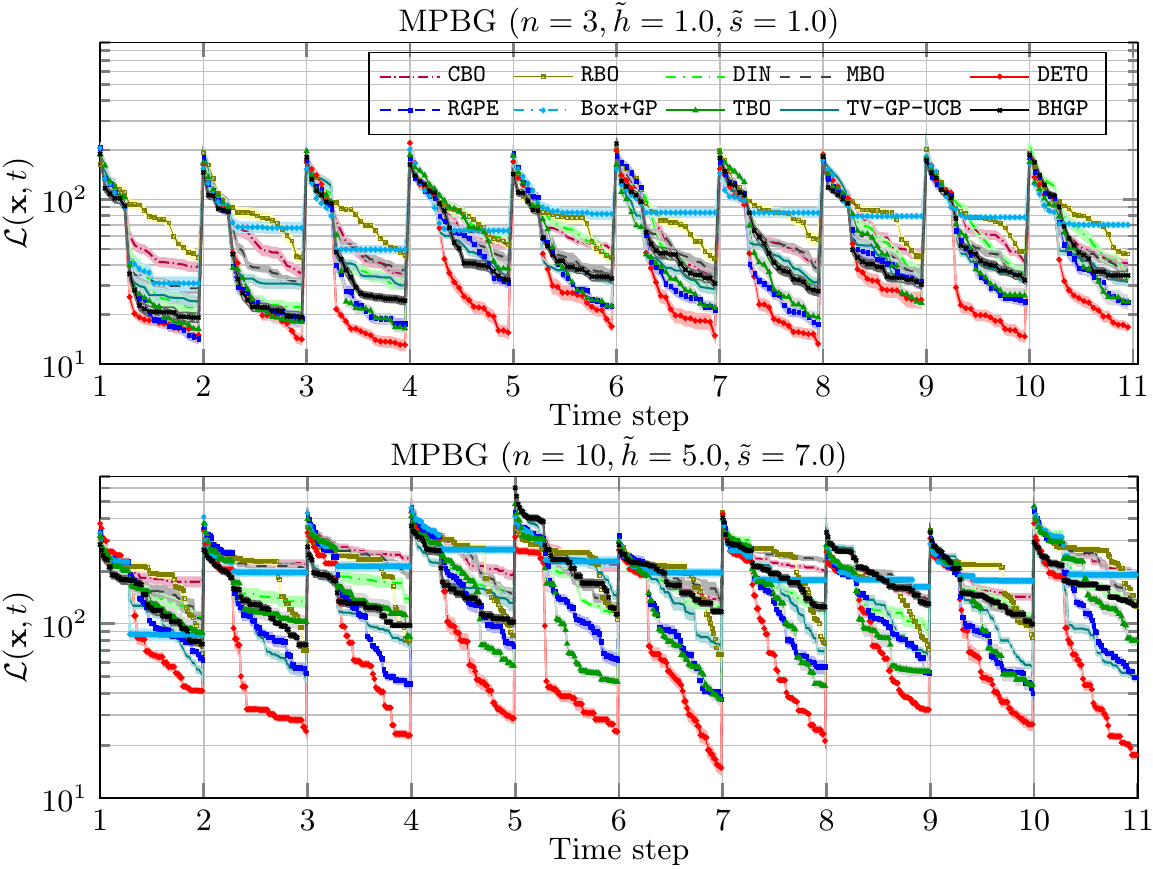}
            \caption{Trajectories of $\mathcal{L}(\mathbf{x},t)$ with confidence bounds over $T=10$ time steps achieved by ten algorithms on two selected MPBG instances.}
            \label{fig:traj_mpbg}
        \end{figure}

    \item As the bar charts of $\rho_\mathrm{c}$ shown in~\pref{fig:rho_c}, we can see that \our\ is still the best algorithm. However, its superiority is not as overwhelming as those evaluated by the $\overline{\epsilon}_\mathrm{f}$ and $\overline{\epsilon}_\mathrm{t}$. The $\rho_c$ obtained by \texttt{TBO} and \texttt{RGPE} are also close to $1.0$, which means they can catch up with \our\ when being allocated with reasonably more FEs. It is interesting to note that \texttt{RBO} is competitive where it even converges faster than some BO variants, especially \texttt{Box+GP}, the worst algorithm in all comparisons.
        \begin{figure}[t!]
            \centering
            \includegraphics[width=\linewidth]{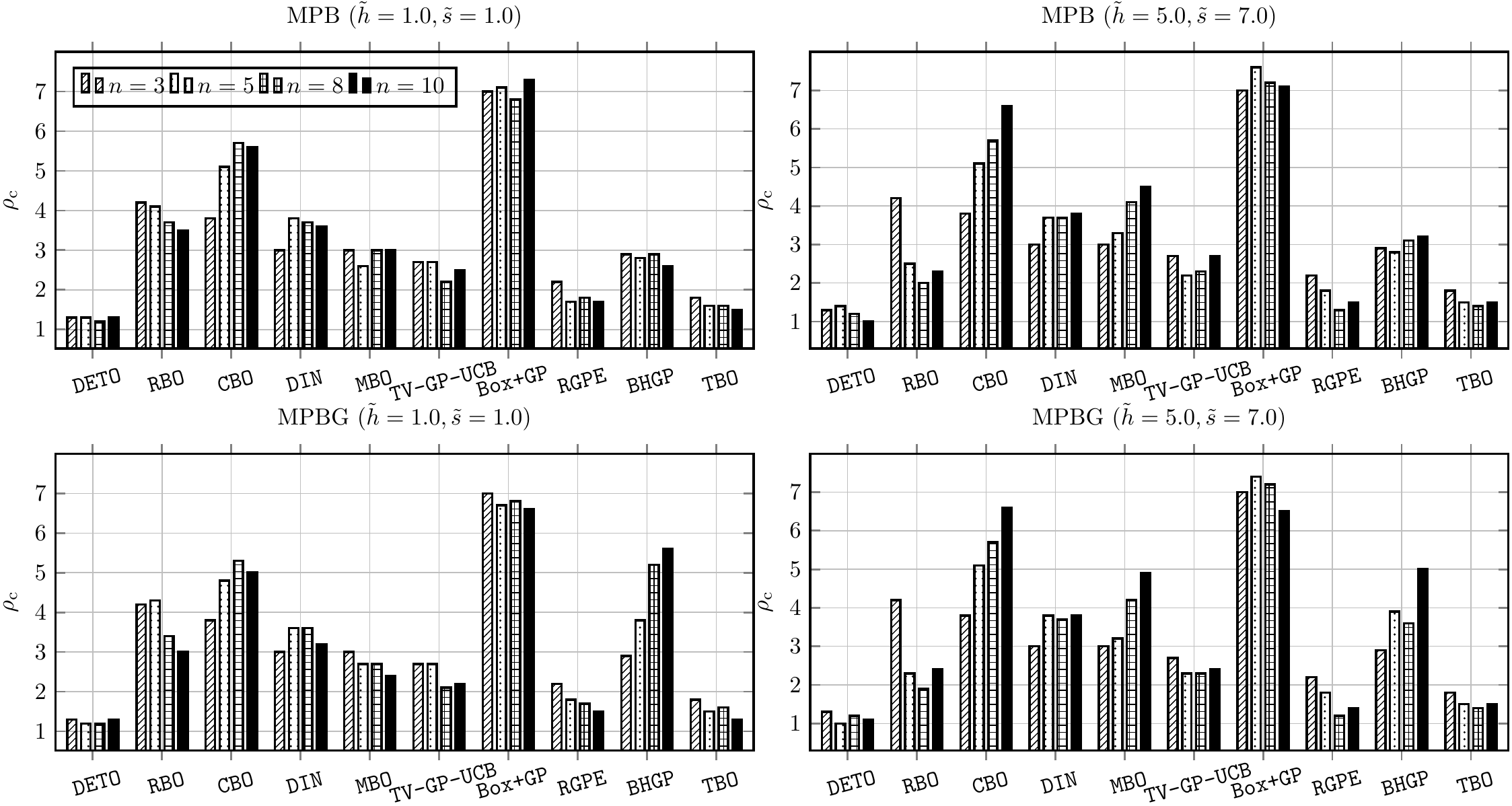}
            \caption{Bar charts of $\rho_\mathrm{c}$ obtained by \our\ and the other nine peer algorithms.}
            \label{fig:rho_c}
        \end{figure}

    \item The bar charts of $\rho_\mathrm{t}$ shown in~\pref{fig:rho_t} demonstrate the effectiveness of the knowledge transfer approaches used in \our, \texttt{TBO}, and \texttt{RGPE}. In particular, \our\ can be one time faster than \texttt{RBO} to achieve its comparable result. On the other hand, the performance of \texttt{CBO} is worse than \texttt{RBO}. This implies that simply feeding all historical data for surrogate modeling can be detrimental to the data-driven optimization. In addition, the knowledge transfer approaches used in \texttt{DIN}, \texttt{MBO}, and \texttt{BHGP} can be negative when handling DOPs with large changes~\cite{LiDY18,CaoKWLLK15,LiWKC13,LiKWCR12,LiKD15,LiDZZ17,CaoKWL14,CaoKWL12,LiKWTM13,LiFK11,LiK14,Li19,LiZLZL09,BillingsleyLMMG19,LiZZL09,ZouJYZZL19,LiNGY22}.
        \begin{figure}[t!]
            \centering
            \includegraphics[width=\linewidth]{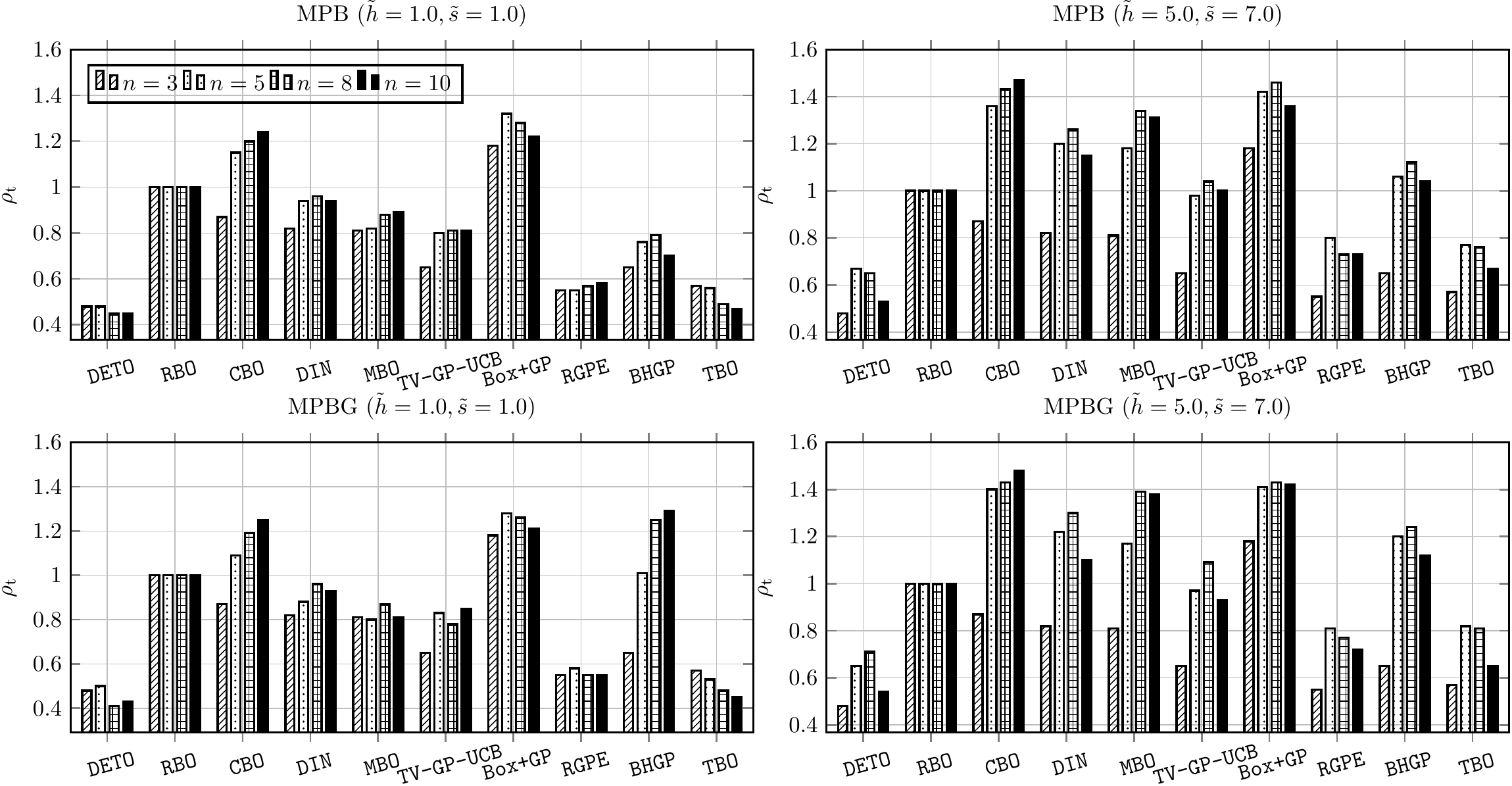}
            \caption{Bar charts of $\rho_\mathrm{t}$ obtained by \our\ and the other nine peer algorithms.}
            \label{fig:rho_t}
        \end{figure}

\end{itemize}

\subsection{Ablation Study}
\label{sec:rq35}

In this subsection, we investigate the effectiveness of the four building blocks of \our\ as introduced in Sections~\ref{sec:mogp} to~\ref{sec:hea}, respectively.

\subsubsection{Methods}
\label{sec:methods_rq36}

To address RQ3 to RQ4, we design the following four variants of \our\ in this ablation study.
\begin{itemize}
    \item To address the RQ3, we develop a variant, denoted as \texttt{DETO-v1}, that replaces the HMOGP with the conventional LMC in the surrogate modeling step. 

    \item To address the RQ4, we develop the following three variants with different source selection mechanisms.
        \begin{itemize}
            \item\texttt{DETO-v2}: it selects the $k$ most recent sources, i.e., $\{\mathcal{D}^i\}_{i=t-k}^{t-1}$ where $t>k$, to build the HMOGP.
            \item\texttt{DETO-v3}: it selects $k$ sources whose hyperparameters of the corresponding GPR models closest to $\mathbf{h}^t$. In other words, the selected source data are supposed to be most similar to the current time step.
            \item\texttt{DETO-v4}: it randomly selects $k$ sources to serve the surrogate modeling purpose.
        \end{itemize}

    \item To address the RQ5, we develop another variant \texttt{DETO-v5} that replaces the warm starting initialization mechanism by a random initialization.

    \item To address the RQ6, we developed two variants \texttt{DETO-v6} and \texttt{DETO-v7} of which the hybrid DE with local search is replaced by a vanilla DE and a vanilla SGD, respectively.
\end{itemize}

\subsubsection{Results}
\label{sec:results_rq36}

Let us interpret the results according to the methods introduced in~\pref{sec:methods_rq36}
\begin{itemize}
    \item From the comparison results of $\bar{\epsilon}_\mathrm{f}$ and $\bar{\epsilon}_\mathrm{t}$ shown in Tables I and II in the Appendix C of the supplemental document along with the $100\%$ large $A_{12}$ effect size shown in~\pref{fig:a12_rq3}, we can see that a significant performance degradation of \our\ when replacing the HMOGP with the conventional LMC. This is counter-intuitive at the first glance. Because the HMOGP can be treated as an approximation of the LMC with a significantly reduced order of hyperparameters, the LMC was supposed to be more expressive and have a more powerful modeling capability than the HMOGP. We attribute the inferior performance of \texttt{DETO-v1} to the underfitting problem discussed in~\pref{sec:mogp}. Let us use an illustrative example of a MPB problem with $n=1$ shown in~\pref{fig:example_rq3} to elaborate this issue. As shown in this figure, both the LMC and the HMOGP work well when $t=2$, i.e., the surrogate model is built by two data sets. However, the LMC becomes clearly underfit when $t=5$ with a surprisingly high uncertainty estimation. It is even worse when $t=10$. As discussed in~\pref{sec:mogp}, the hyperparameters of the LMC surges to $1,000$ when $t=10$ whereas the number of training data is way smaller. In contrast, the HMOGP performs consistently well even when $t=10$ as its number of hyperparameters is merely $10$, which is manageable~\cite{XuLA22,PruvostDLL020,WuLLL22}.
        \begin{figure}[t!]
            \centering
            \includegraphics[width=.6\linewidth]{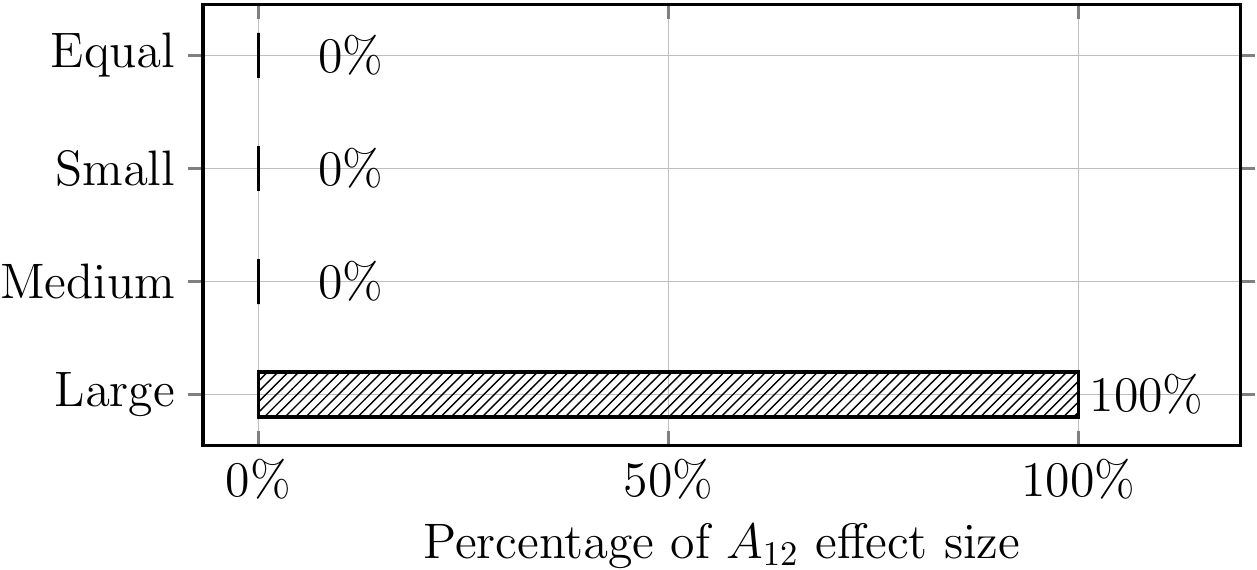}
            \caption{Percentage of the large, medium, small, and equal $A_{12}$ effect size of $\bar{\epsilon}_\mathrm{f}$ and $\bar{\epsilon}_\mathrm{t}$ when comparing \our\ with \texttt{DETO-v1}.}
            \label{fig:a12_rq3}
        \end{figure}

        \begin{figure}[t!]
            \centering
            \includegraphics[width=\linewidth]{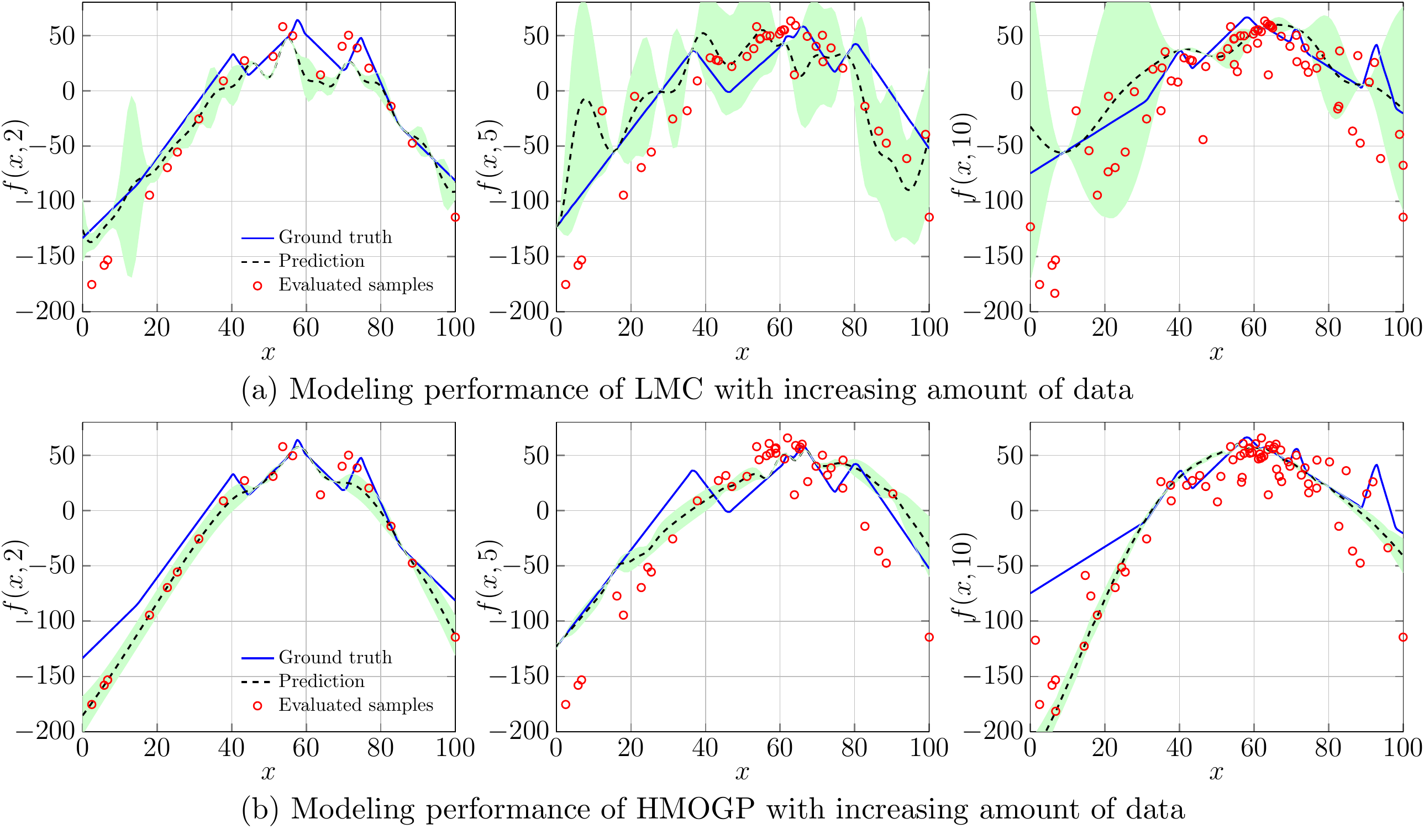}
            \caption{An illustrative example of underfitting problem of using the LMC versus the HMOGP for surrogate modeling of MPB problems.}
            \label{fig:example_rq3}
        \end{figure}

    \item As the comparison results of $\bar{\epsilon}_\mathrm{f}$ and $\bar{\epsilon}_\mathrm{t}$ shown in Tables III and IV in the Appendix C of the supplemental document along with the $A_{12}$ effect size shown in~\pref{fig:a12_rq4}, we confirm that our proposed algorithm benefits from the adaptive source data selection mechanism. It is counter-intuitive to see that \texttt{DETO-v3}, which picks up the most similar $k$ sources, is the worst variant. It is even outperformed by \texttt{DETO-v4} which randomly selects $k$ sources for surrogate modeling. This can be explained as the data collected during the initialization of each time step is not sufficient to build a reasonably good surrogate model. Thus, choosing the similar source tasks can be misleading accordingly. It is interesting to note that the performance of \our\ is close to that of \texttt{DETO-v2} when $\tilde{h}=1.0$ and $\tilde{s}=1.0$, i.e., the dynamic problems with a small change. This can be explained as the landscape of such problem is close to its recent time steps, it makes sense to leverage the data collected from therein~\cite{WuLKZ20,LiLDMY20,LiCSY19,WuKJLZ17,LiLLM21,LaiL021,ShanL21,WangYLK21,LiXCT20,LiuLC20,LiX0WT20,CaoWKL11,KumarBCLB18,LiZ19,LiuLC19,GaoNL19,LiXT19,LiDZ15,LiDAY17,WuKZLWL15}.
        \begin{figure}[t!]
            \centering
            \includegraphics[width=.6\linewidth]{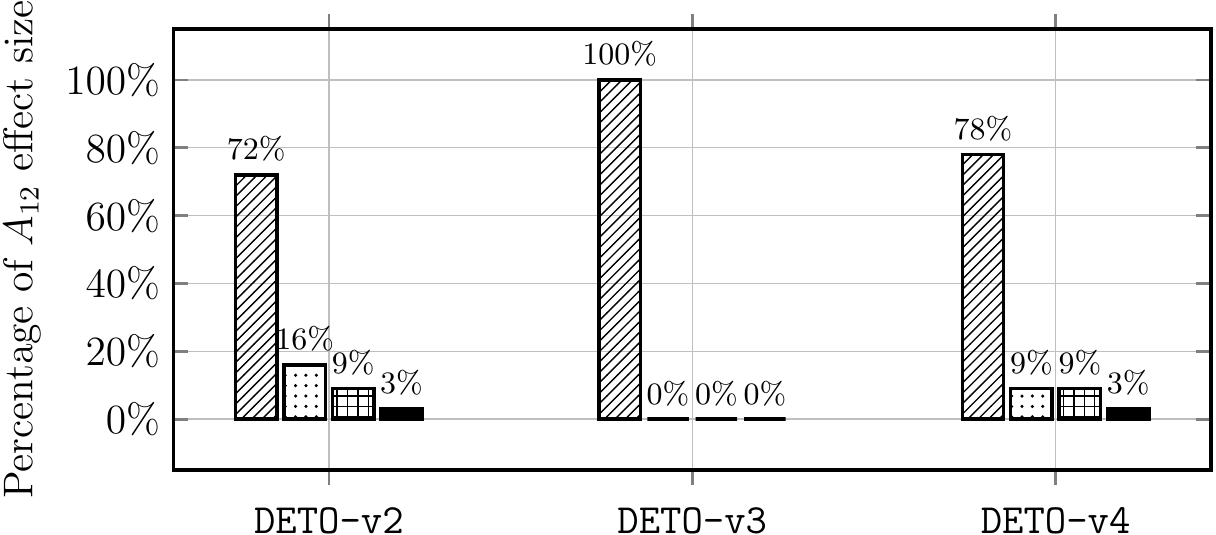}
            \caption{Percentage of the large, medium, small, and equal $A_{12}$ effect size of $\bar{\epsilon}_\mathrm{f}$ and $\bar{\epsilon}_\mathrm{t}$ when comparing \our\ with \texttt{DETO-v2}, \texttt{DETO-v3}, and \texttt{DETO-v4}.}
            \label{fig:a12_rq4}
        \end{figure}

    \item From the comparison results of $\bar{\epsilon}_\mathrm{f}$ and $\bar{\epsilon}_\mathrm{t}$ shown in Tables V and VI in the Appendix C of the supplemental document along with the $A_{12}$ effect size shown in~\pref{fig:a12_rq5}, we can see the effectiveness of the warm starting initialization mechanism introduced in~\pref{sec:initialization}. However, it is also worth noting that the overwhelmingly better performance is diminishing when $n=10$. This can be explained as the ineffectiveness of the surrogate modeling in a high-dimensional scenario with a strictly limited amount of training data. In this case, the local optima estimated from a less reliable model can be misleading~\cite{RuanLDL20,XuLA21,Chen021,WilliamsLM21}.
    \begin{figure}[t!]
            \centering
            \includegraphics[width=.6\linewidth]{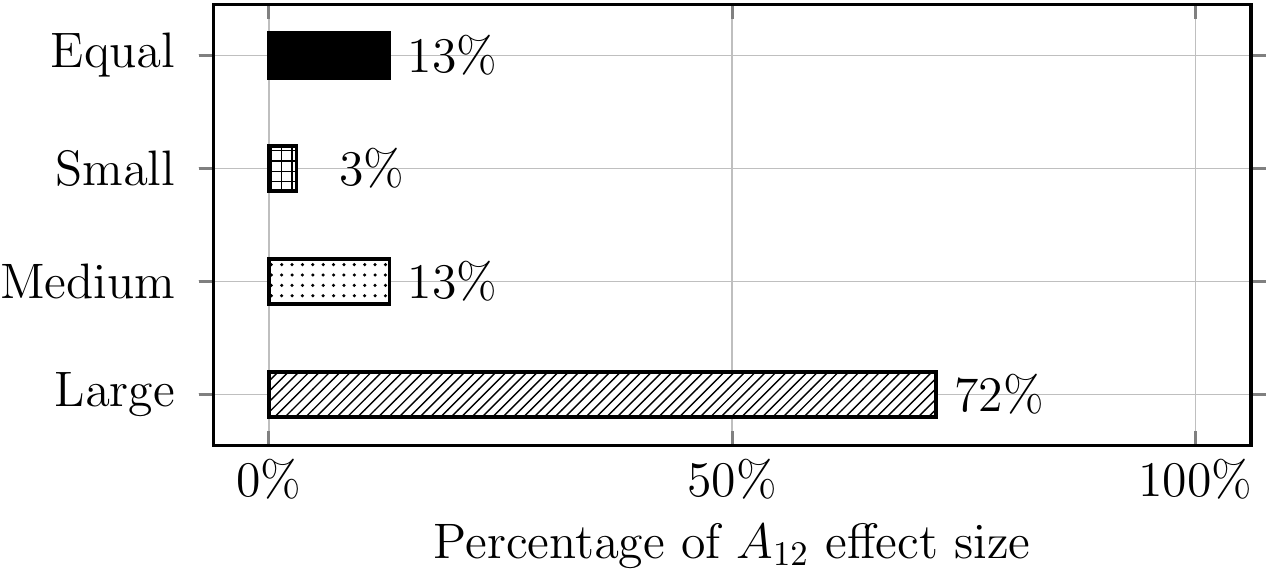}
            \caption{Percentage of the large, medium, small, and equal $A_{12}$ effect size of $\bar{\epsilon}_\mathrm{f}$ and $\bar{\epsilon}_\mathrm{t}$ when comparing \our\ with \texttt{DETO-v5}.}
            \label{fig:a12_rq5}
        \end{figure}

    \item From the comparison results of $\bar{\epsilon}_\mathrm{f}$ and $\bar{\epsilon}_\mathrm{t}$ shown in Tables VII and VIII in the Appendix C of the supplemental document along with the $A_{12}$ effect size shown in~\pref{fig:a12_rq4}, we can see that our proposed hybrid DE with local search is the most effective optimizer for the acquisition function. Due to the non-convex property, neither vanilla DE nor SGD works well as both of them have a risk of being stuck at some local optima.
    \begin{figure}[t!]
            \centering
            \includegraphics[width=.6\linewidth]{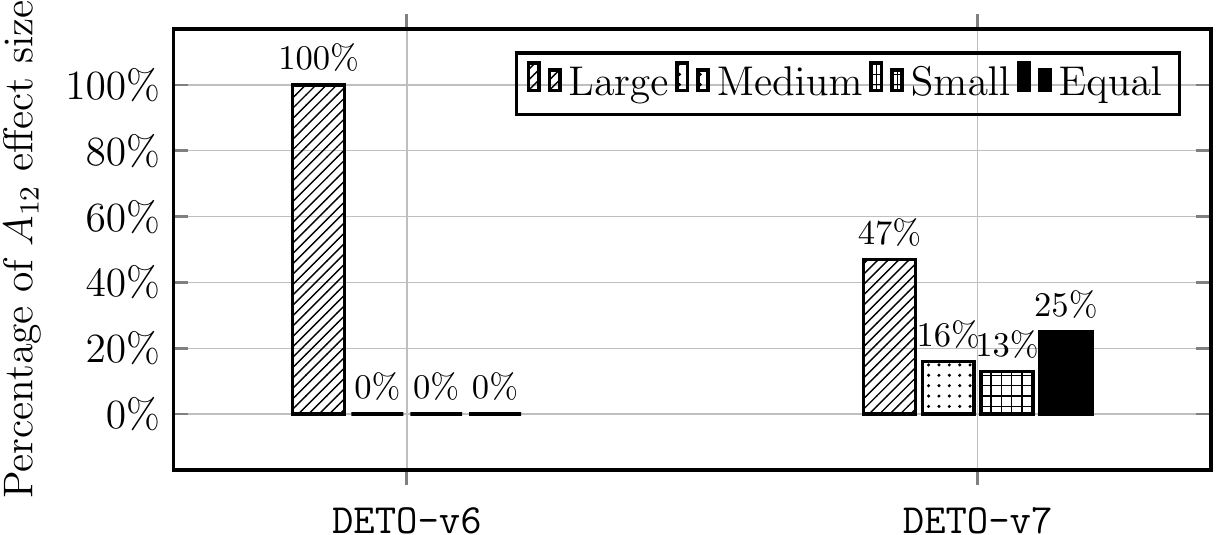}
            \caption{Percentage of the large, medium, small, and equal $A_{12}$ effect size of $\bar{\epsilon}_\mathrm{f}$ and $\bar{\epsilon}_\mathrm{t}$ when comparing \our\ with \texttt{DETO-v6} and \texttt{DETO-v7}.}
            \label{fig:a12_rq6}
        \end{figure}
\end{itemize}

\subsection{Real-World Dynamic Optimization}
\label{sec:rq7}

At the end, we validate the performance of \our\ and the other nine peer algorithms on a real-world DOP of which the unknown objective function evolves over time. We consider here an automated machine learning task for handwritten digit recognition of the famous MNIST dataset~\cite{LarochelleECBB07}. As an illustrative example shown in~\pref{fig:mnist_rotation}, the digits are periodically rotated with time. Therefore, the hyperparameters associated with the baseline classifier, a multi-layer perceptron (MLP) in particular, need to be adjusted accordingly. The experimental settings are introduced as follows.
\begin{itemize}
    \item In our experiments, we create $11$ progressively changing datasets with $60,000$ handwritten digit images for training and another $10,000$ ones for testing. Each image has $28\times 28$ pixels. In particular, to create dynamically changing scenarios, the images are rotated from $0^\circ$ to $360^\circ$ with a step size of $36^\circ$.
        \begin{figure}[t!]
            \centering
            \includegraphics[width=.6\linewidth]{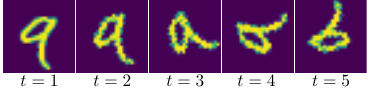}
            \caption{Illustrative example of the MNIST dataset with rotation at different time steps used in our experiments (detailed examples can be found in Fig. 5 of the Appendix D of the supplemental document).}
            \label{fig:mnist_rotation}
        \end{figure}

    \item We consider tuning four hyperparameters associated with the MLP outlined as below.
        \begin{itemize}
            \item The number of neurons in the each hidden layer $n^h\in \{16,32,48,64,80,96,112,128\}$.
            \item The learning rate: $\alpha\in [10^{-6}, 1.0]$.
            \item The momentum factor in SGD: $\beta \in [0.0,1.0]$.
            \item The weight for activation functions for the hidden layer $w \in [0.0,1.0]$.
        \end{itemize}
\end{itemize}

\begin{figure}[t!]
    \centering
    \includegraphics[width=.6\linewidth]{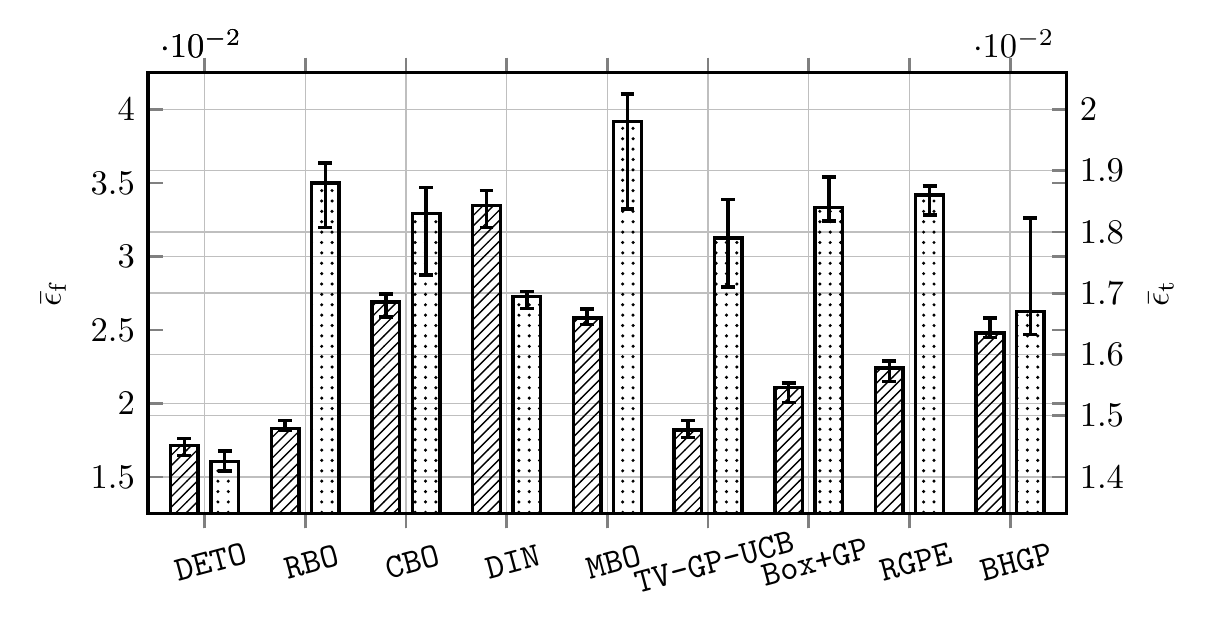}
    \caption{Bar charts with error bars of $\bar{\epsilon}_\mathrm{f}$ and $\bar{\epsilon}_\mathrm{t}$ achieved by \our\ against the other nine peer algorithms on the real-world case study.}
    \label{fig:mnist_box}
\end{figure}

From the comparison results shown in~\pref{fig:mnist_box}, it is clear to see that our proposed \our\ consistently outperforms the other peer algorithms on both $\overline{\epsilon}_\mathrm{f}$ and $\overline{\epsilon}_\mathrm{t}$ metrics. It is interesting to note that \texttt{RBO} is the second best algorithm on the $\overline{\epsilon}_\mathrm{f}$ metric whereas it becomes the second worst one when considering the $\overline{\epsilon}_\mathrm{t}$ metric. This observation implies that the transfer learning approaches used in other peer algorithms might lead to negative effect at the early stage of the hyperparameter optimization at each time step. Whereas a simple restarting from scratch is more robust at the early stage while its performance might be stagnated afterwards~\cite{Williams0M22,ZhouLM22,LiLL22,Zhou0M22}.


\section{Conclusions}
\label{sec:conclusions}

This paper develops a simple but effective data-driven transfer optimization framework that empowers data-driven evolutionary optimization to tackle expensive black-box optimization problems under dynamically changing environments. The main crux of our proposed \our\ is a hierarchical MOGP as the surrogate model to implement a cost-effective knowledge transfer across the data collected from different time steps. In addition, an adaptive source task selection mechanism and a bespoke warm staring initialization strategy further enable a better knowledge transfer. In future, our proposed \our\ framework can be extended from the following aspects.
\begin{itemize}
    \item First, there has been a wealth of works on transfer learning in the literature~\cite{YangZDP20}. In principle, any prevalent transfer learning approach can be applied in \our\ in a plug-in manner. However, as discussed in~\pref{sec:rq35}, extra attention is required to mitigate the underfitting problems caused by the extremely limited computational budget allocated in DOPs.
    \item Meta-learning have been an emerging approach to learn new concepts and skills fast with a few training examples by a learning to learn paradigm~\cite{HospedalesAMS22}. It is promising to unleash the potential of meta-learning to tackle expensive DOPs under our proposed \our\ framework.
    \item Last but not the least, existing transfer optimization approaches~\cite{TanFJ21,ChenLTL22} are hardly scalable to many tasks. However, real-world systems are usually designed to tackle a large number of (even infinitely many) problems over their lifetime. As one of our next ambitions, it is interesting to develop a life-long optimization paradigm that continuously learn knowledge from ongoing optimization practices and autonomously select useful information to new and unseen tasks in an open-ended environment.
\end{itemize}

\section*{Acknowledgment}
This work was supported by UKRI Future Leaders Fellowship (MR/S017062/1), EPSRC (2404317), NSFC (62076056), Royal Society (IES/R2/212077) and Amazon Research Award.

\bibliographystyle{IEEEtran}
\bibliography{IEEEabrv,tbo2}

\end{document}